\PassOptionsToPackage{table}{xcolor}
\documentclass[ASNA,twocolumn]{USG}
\usepackage{anyfontsize}
\usepackage{colortbl}
\usepackage{longtable}
\usepackage{booktabs}
\usepackage{arydshln}
\usepackage{array}
\definecolor{myred}{rgb}{1.0, 0.0, 0.0}
\definecolor{myblue}{rgb}{0.0, 0.0, 1.0}
\definecolor{lightgrayline}{RGB}{200,200,200}

\graphicspath{{./images/}{./figures/}}
\newcommand{\beginsupplement}{%
        \setcounter{figure}{0}
        \renewcommand{\thefigure}{S\arabic{figure}}
        \renewcommand{\theHfigure}{S\arabic{figure}}}

\makeatletter
\providecommand{\@setmarks}{}
\makeatother
\articletype{RESEARCH ARTICLE}
\journal{Advanced Science}
\volume{}
\copyyear{2026}
\startpage{1}
\articledoi{}

\begin{document}

\title{{\mdseries \textbf{MIRCID}: Inferred Hub-\textbf{miR}NAs Drive \textbf{C}ross-Task \textbf{I}mprovements in \textbf{D}rug Mechanistic Modeling}}

\author[1,2,3]{Xin Cao}[][\textsuperscript{\textdagger}]
\author[1,3]{Yigang Chen}[][\textsuperscript{\textdagger}]
\author[1,3]{Jiatong Xu}[][\textsuperscript{\textdagger}]
\author[1,3]{Ziyue Zhang}[][\textsuperscript{\textdagger}]
\author[1]{Xiang Cheng}
\author[1]{Shenyu Wang}
\author[1]{Yangyi Zhang}
\author[1,3]{Xiaoxuan Cai}
\author[1,3]{Shidong Cui}
\author[1,3]{Zihao Zhu}
\author[1,3]{Xiang Ji}
\author[1,3,4]{Hsi-Yuan Huang}
\author[1,3,4]{Yang-Chi-Dung Lin}
\author[2,4,5]{Tianwei Yu}
\author[1,3,4,6]{Hsien-Da Huang}[][\textsuperscript{*}]

\authormark{CAO \textsc{et al.}}
\titlemark{MIRCID}

\address[1]{School of Medicine, The Chinese University of Hong Kong, Shenzhen, Longgang District, Shenzhen, Guangdong 518172, China}
\address[2]{School of Data Science, The Chinese University of Hong Kong, Shenzhen, Longgang District, Shenzhen, Guangdong 518172, China}
\address[3]{Warshel Institute for Computational Biology, School of Medicine, The Chinese University of Hong Kong, Shenzhen, Longgang District, Shenzhen, Guangdong 518172, China}
\address[4]{Guangdong Provincial Key Laboratory of Digital Biology and Drug Development, The Chinese University of Hong Kong, Shenzhen, Longgang District, Shenzhen, Guangdong 518172, China}
\address[5]{Department of Biostatistics and Bioinformatics, Rollins School of Public Health, Emory University, Atlanta, GA 30322, USA}
\address[6]{Department of Endocrinology, Key Laboratory of Endocrinology of National Health Commission, Peking Union Medical College Hospital, Chinese Academy of Medical Sciences \& Peking Union Medical College, Beijing, 100730, P.R. China}

\historydates{{\FIfont\textsuperscript{\textdagger}These authors contributed equally to this work.\quad corresponding authors: Tianwei Yu (yutianwei@cuhk.edu.cn) and Hsien-Da Huang (huanghsienda@cuhk.edu.cn)}}


\abstract[ABSTRACT]{
Drug mechanism-of-action (MoA) modeling commonly relies on perturbational transcriptomes, but matched microRNA (miRNA) measurements are often unavailable. Inferred regulatory features offer a scalable way to reuse these data. Here, we present MIRCID, a framework comparing gene expression with inferred transcription factor (TF) activity and miRNA expression across pathway classification and similarity-based MoA retrieval. HubmiRNet infers 414 pan-cancer hub miRNAs (HubmiRs) from 977 L1000 landmark genes, achieving a Pearson correlation coefficient of 87.72\%; its 1,298-output variant also outperformed SiCmiR on the full-miRNA task (71.21\% versus 67.30\%). In the evaluated comparisons, miRNA augmentation provided more consistent gains than TF activity. Generic embedding controls showed model-dependent utility, while complementarity analyses identified a distinct, partially linearly recoverable representation that retained gene-derived structure. Illustrative rescue cases linked improved classification to biologically plausible miRNA patterns in samples with weak transcriptional signatures. These findings support inferred HubmiRs as a biologically informed recoding of transcriptomic data for perturbational drug modeling, while leaving recovery of measured perturbational miRNA responses to further validation.
}

\flushbottom
\maketitle

\section{Introduction}

Elucidating the mechanism of action (MoA) of small molecules remains a central bottleneck in drug discovery \cite{iorioDiscoveryAnticancer2017}. High-throughput transcriptomic profiling, exemplified by the L1000 assay \cite{subramanianNextGenCMap2017}, has become a standard strategy for capturing cellular responses to chemical perturbation. However, reliance on differential gene expression alone presents both biological and computational limitations. Biologically, endpoint transcriptomic signatures provide only a partial view of cellular regulation, treating the cell largely as a ``black box'' while obscuring the upstream control logic that shapes phenotypic responses. Moreover, drug-induced transcriptional responses are often subtle, transient, or partially silent, such that key targets or pathways may be actively perturbed without producing strong differential expression signals \cite{jiangPathwaySignatures2025}. Computationally, transcriptomic profiles are high-dimensional and noisy, making predictive models prone to overfitting and weakening their ability to generalize across biological contexts \cite{daiHiFIT2025}. As a result, models trained exclusively on endpoint gene expression often show limited robustness when applied to mechanistic inference or therapeutic prioritization tasks \cite{iorioDiscoveryAnticancer2017}.

\begin{figure*}[!b]
    \begin{center}
    \includegraphics[width=0.80\textwidth]{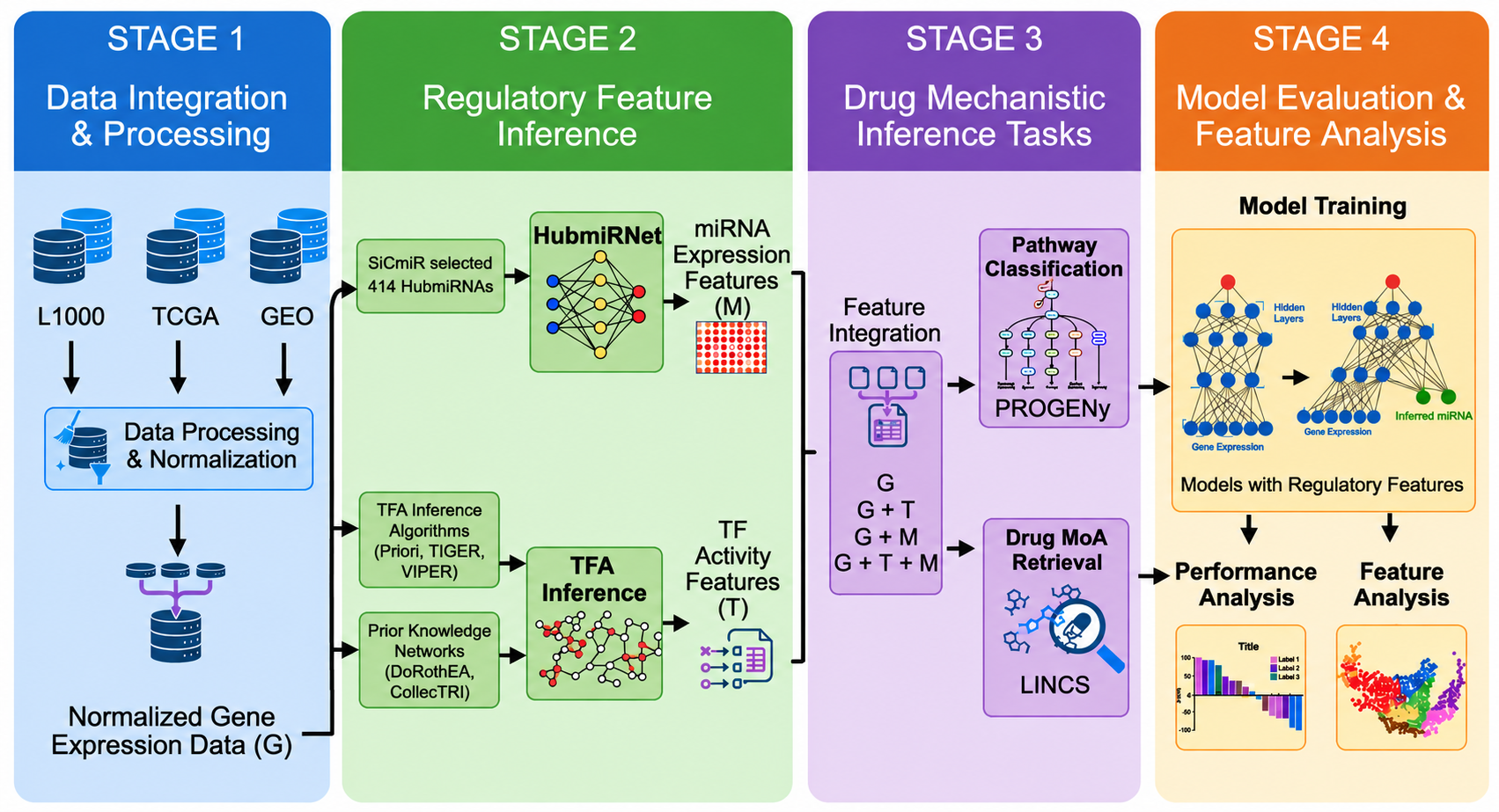}
    \end{center}
    \caption[Workflow of the proposed framework for integrating regulatory layers into drug mechanistic modeling.]{\textbf{Workflow of the proposed framework for integrating regulatory layers into drug mechanistic modeling.} The workflow comprises four stages: data integration; miRNA/TF regulatory feature inference (including HubmiRNet); systematic comparison of regulatory feature layers across downstream perturbational drug modeling tasks; and model training with performance and feature-contribution evaluation.}
    \label{fig:workflow}
\end{figure*}

\begin{figure*}[!t]
    \begin{center}
    \includegraphics[width=0.76\textwidth]{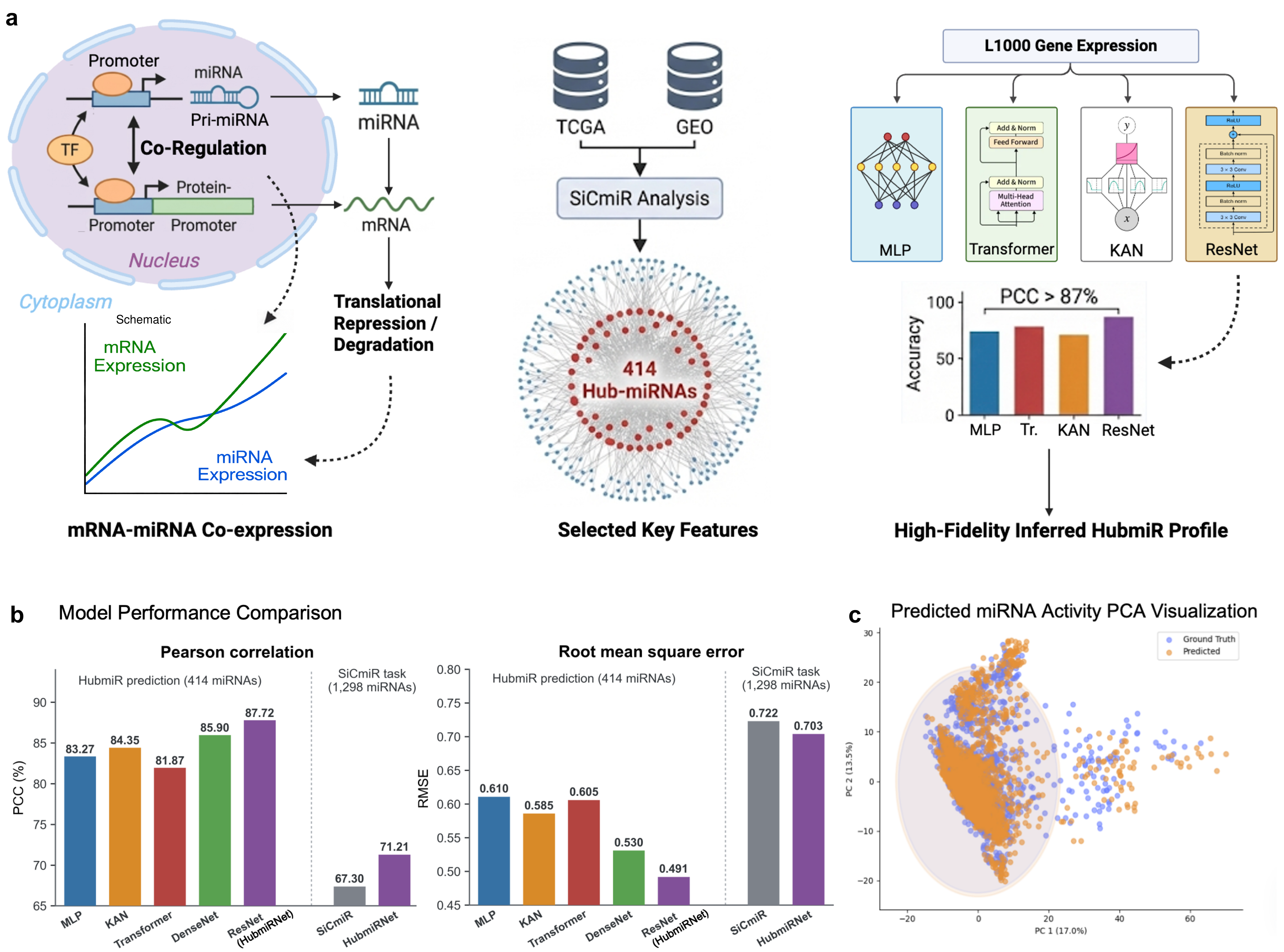}
    \end{center}
    \caption[HubmiRNet enables accurate inference of miRNA expression from mRNA profiles.]{\textbf{HubmiRNet enables accurate inference of miRNA expression from mRNA profiles.} \textbf{a,} mRNA--miRNA co-expression basis and model development. Building on the 414 hub miRNAs identified in SiCmiR \cite{caiSiCmiRAtlas2025}, we benchmarked multiple architectures for inferring miRNA expression from L1000 landmark-gene inputs. \textbf{b,} Pearson correlation coefficient (PCC) and root mean square error (RMSE) for the 414-HubmiR architecture benchmark and for a matched comparison between SiCmiR and a full-output HubmiRNet variant on the original 1,298-miRNA SiCmiR task. \textbf{c,} Principal component analysis of predicted and observed 414-HubmiR profiles, showing overlap in their global variance structure.}
    \label{fig:hubmir}
\end{figure*}

\begin{figure*}[!t]
    \begin{center}
    \includegraphics[width=0.90\textwidth]{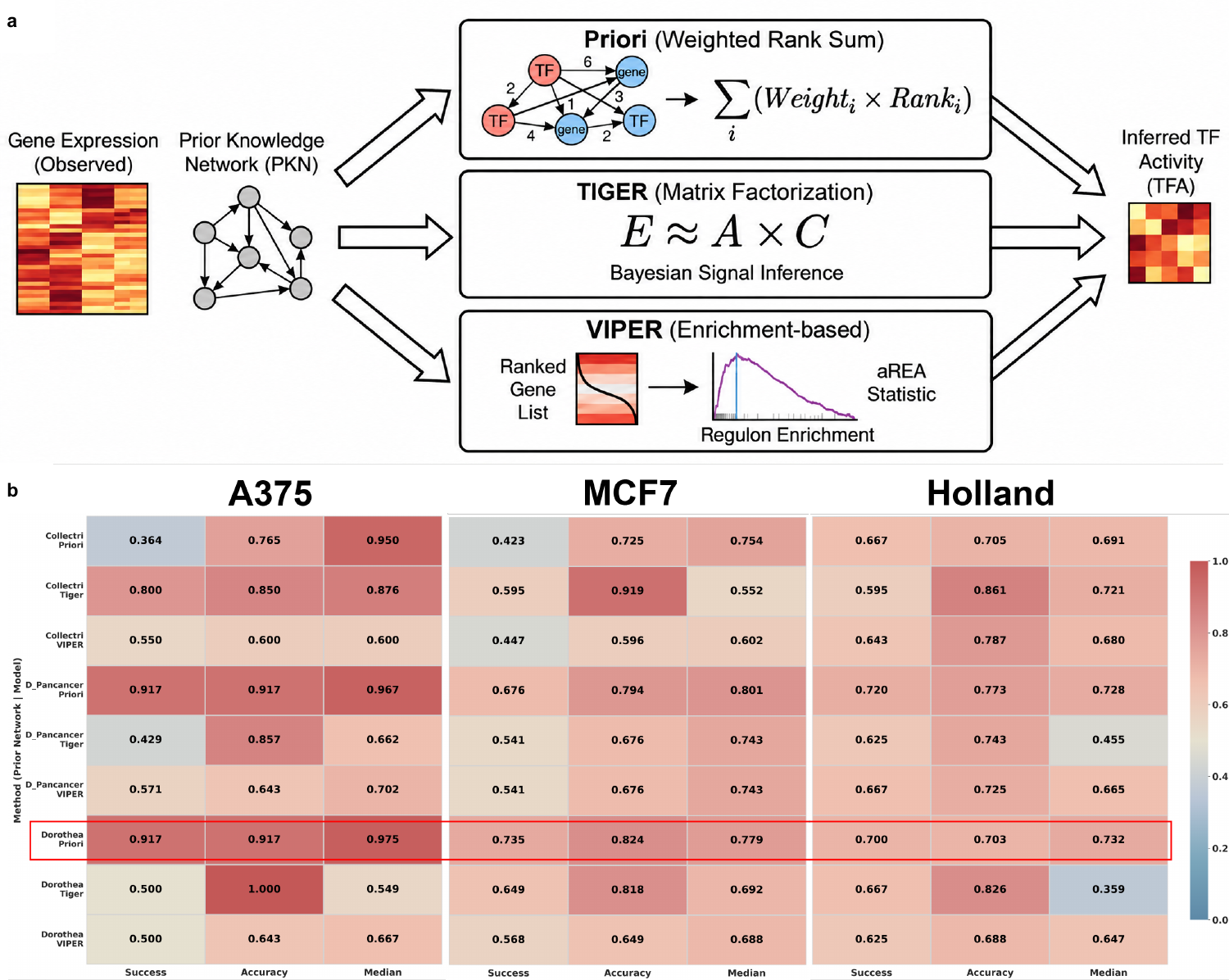}
    \end{center}
    \caption[Systematic benchmarking identifies a robust TF activity inference configuration.]{\textbf{Systematic benchmarking identifies a robust TF activity inference configuration.} \textbf{a,} Benchmarking workflow for TF activity (TFA) inference: gene expression profiles are combined with alternative prior knowledge networks (PKNs) to evaluate three representative TFA algorithms under a unified pipeline. \textbf{b,} Summary of benchmarking results across three perturbation datasets, showing that the DoRothEA+Priori configuration achieves the best or second-best performance for most evaluation metrics.}
    \label{fig:tfa}
\end{figure*}

\begin{figure*}[!t]
    \noindent\hspace*{0.03\textwidth}\includegraphics[width=0.98\textwidth]{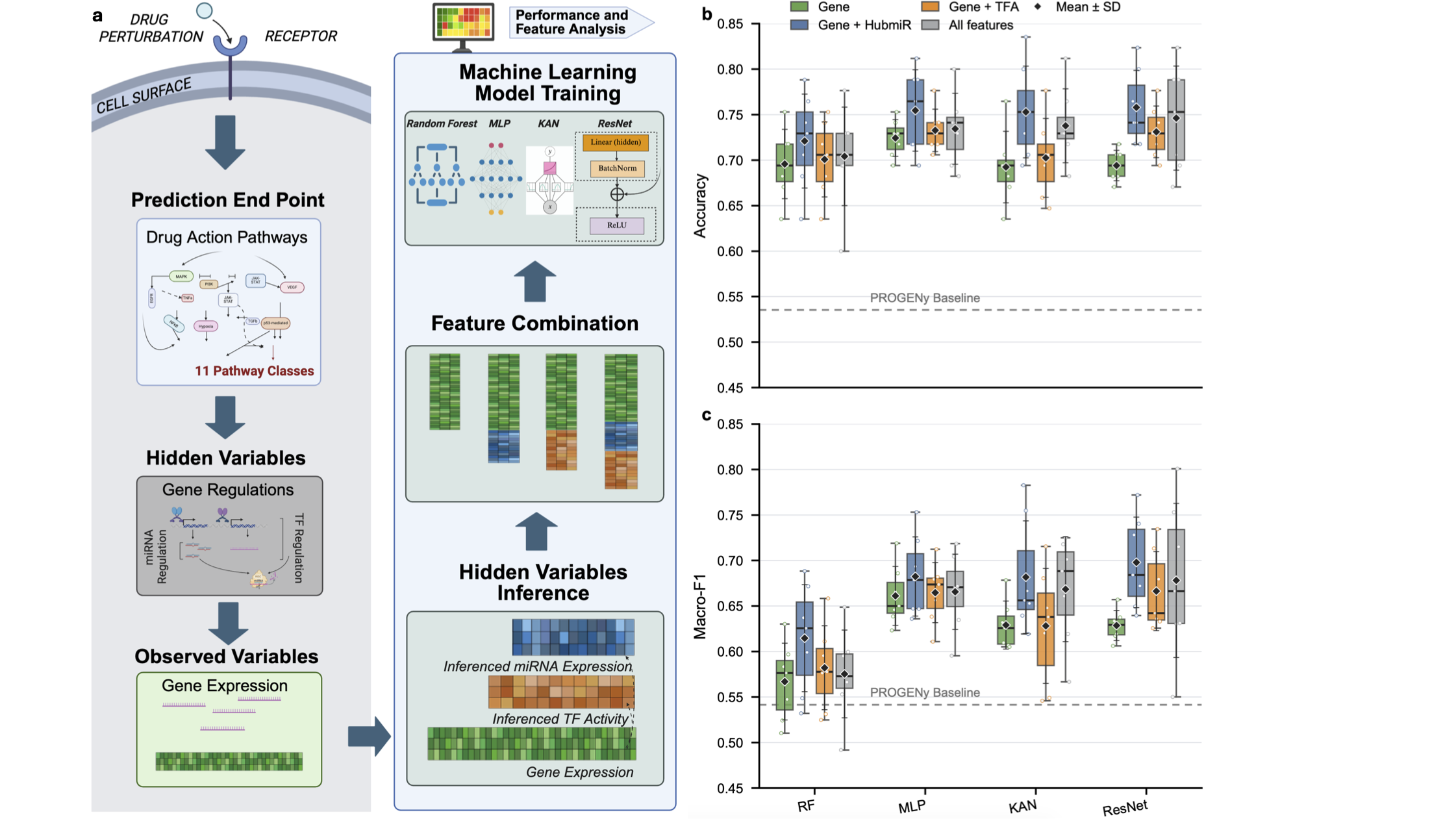}
    \caption[miRNA features consistently improve drug pathway inference performance across models.]{\textbf{miRNA features consistently improve drug pathway inference performance across models.} \textbf{a,} Pathway-classification workflow using Gene alone or Gene combined with inferred TF activity (TFA), inferred HubmiR expression, or both. \textbf{b,c,} Held-out accuracy (\textbf{b}) and macro-F1 (\textbf{c}) for random forest (RF), multilayer perceptron (MLP), Kolmogorov--Arnold network (KAN), and residual network (ResNet) classifiers. Points denote individual held-out runs; boxes show the median and interquartile range, whiskers the minimum and maximum, and diamonds with bars the mean $\pm$ s.d. Dashed lines show the mean gene-only PROGENy linear-model performance across 7 splits.}
    \label{fig:path_performance}
\end{figure*}

\begin{figure*}[!t]
    \begin{center}
    \includegraphics[width=0.84\textwidth]{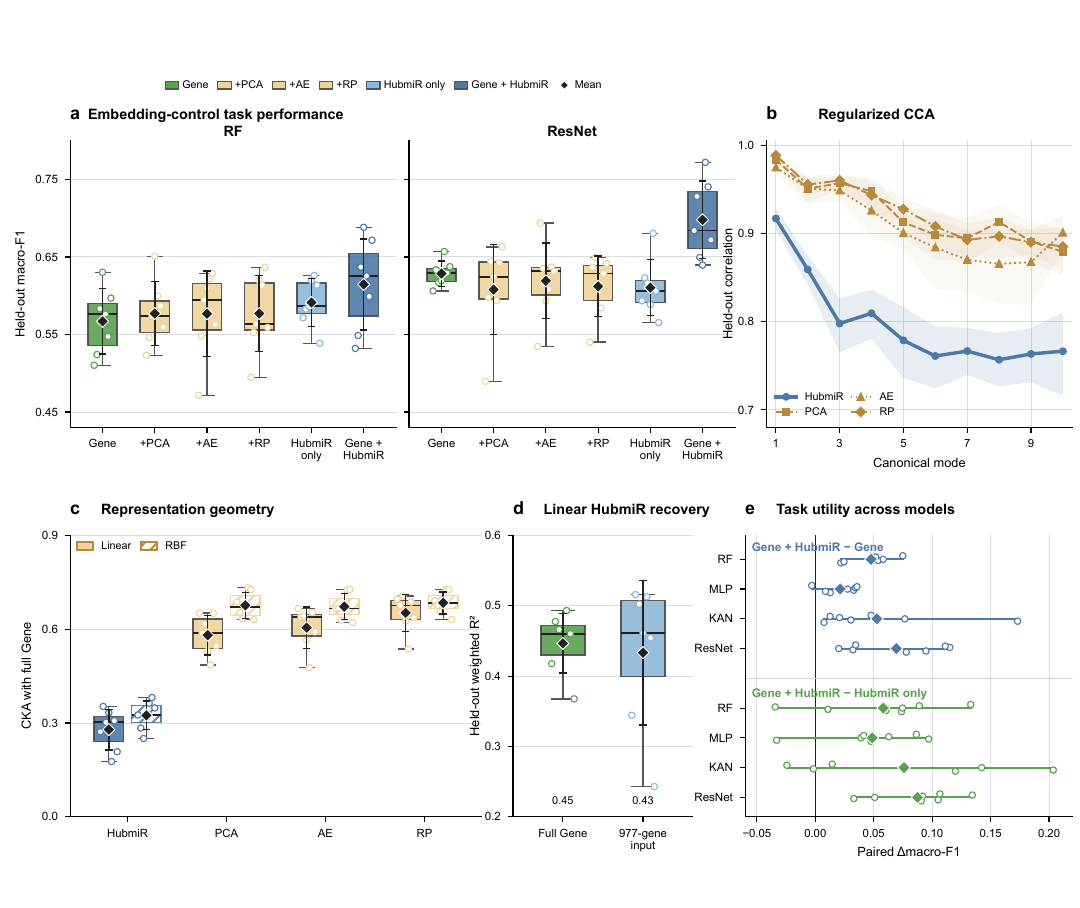}
    \end{center}
    \caption[Generic embedding controls and complementary geometry of the inferred HubmiR representation.]{\textbf{Generic embedding controls and complementary geometry of the inferred HubmiR representation.} \textbf{a,} Held-out macro-F1 for RF and ResNet using Gene, Gene plus 414-dimensional principal-component analysis (PCA), autoencoder (AE), or random projection (RP), HubmiR alone, or Gene+HubmiR. Controls used the same 977 landmark-gene inputs; RP is averaged over five projections. \textbf{b,} Regularized canonical correlation analysis (CCA) between Gene and each representation across the first ten held-out modes (mean and bootstrap 95\% CI; 20 splits). \textbf{c,} Debiased linear and radial-basis-function centred-kernel alignment (CKA) with Gene (seven evenly spaced display splits from 20). \textbf{d,} Held-out variance-weighted $R^2$ from cross-validated multi-output ridge regression predicting HubmiR from full Gene or the 977-gene input for the same seven splits. \textbf{e,} Paired change in held-out macro-F1 after adding HubmiR to Gene or Gene to HubmiR across RF, MLP, KAN, and ResNet. In \textbf{a}, boxes show the median and interquartile range and whiskers the observed range; in \textbf{e}, diamonds denote means and lines the observed range.}
    \label{fig:feature_analysis}
\end{figure*}

\begin{figure*}[!b]
    \noindent\hspace*{0.12\textwidth}\includegraphics[width=0.7\textwidth]{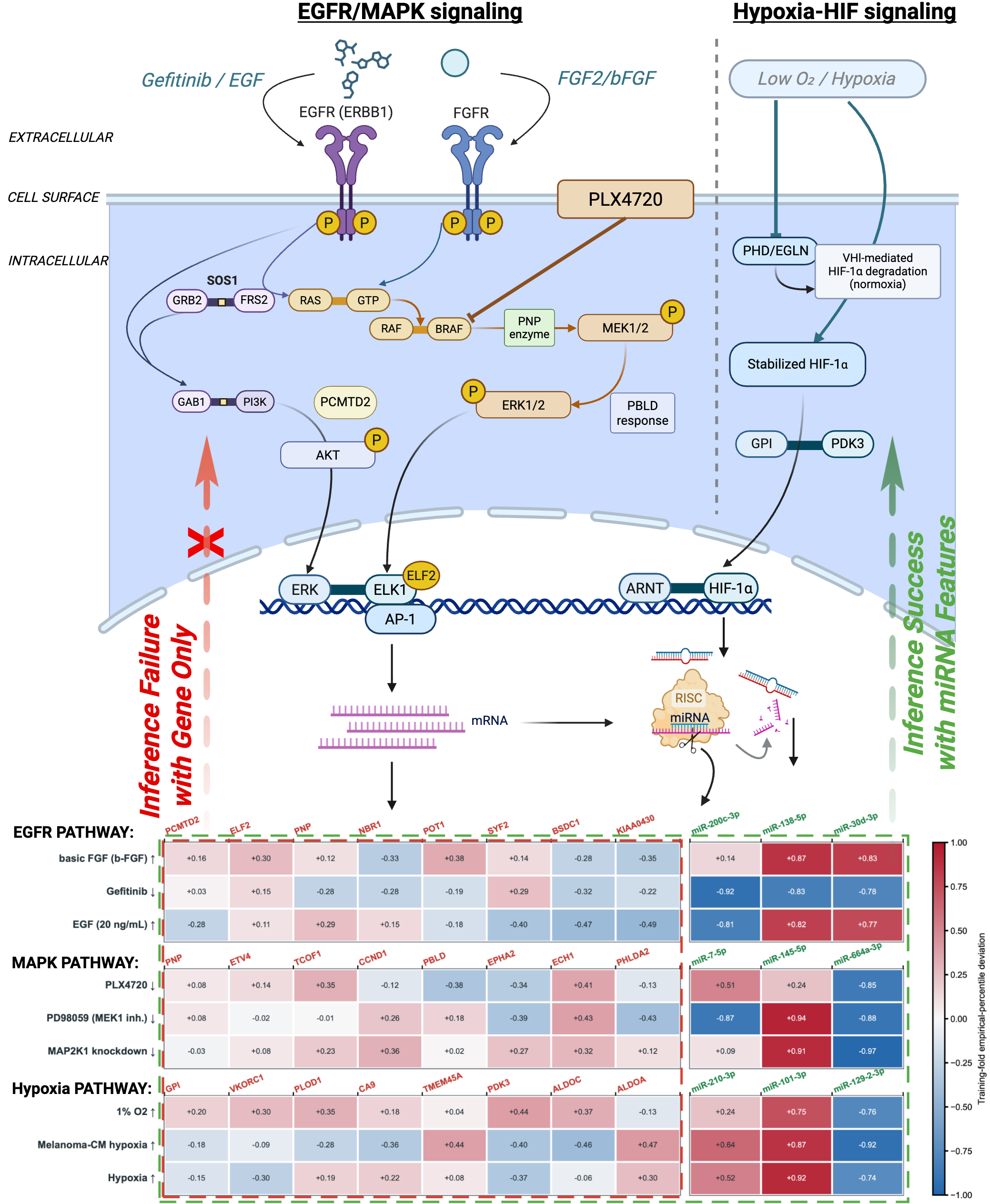}
    \caption[Inferred HubmiR features rescue pathway classification across EGFR, MAPK, and hypoxia perturbations.]{\textbf{Inferred HubmiR features rescue pathway classification across EGFR, MAPK, and hypoxia perturbations.} The schematic summarizes the represented signaling contexts. Heatmaps show nine strict-rescue cases (three per pathway) in which Gene-only ResNet was incorrect and Gene+HubmiR was correct on the same held-out split. Each case displays eight pathway-associated PROGENy genes and three pathway-related inferred HubmiRs as training-fold empirical-percentile deviations ($-1$ to $+1$; 0, training-fold median). Arrows denote activating or inhibiting perturbations; red and green frames identify the displayed gene and HubmiR features, respectively. When pathway-related gene signals were weak, the corresponding pathway-related HubmiRs showed larger deviations, supporting rescue by complementary miRNA features.}
    \label{fig:path_case}
\end{figure*}

Drug responses propagate through transcriptional and post-transcriptional control, involving both transcription factors (TFs) and microRNAs (miRNAs) \cite{bartelMicroRNAs2009,jonasIzaurralde2015}. MiRNA-containing feedback and feedforward circuits coordinate transcriptional programs with target regulation, providing a biological rationale for considering miRNAs when characterizing network state \cite{Tsang2007NetworkMotifs}. Measured miRNA profiles also reflect tumour lineage and differentiation state, demonstrating their relevance to cellular context \cite{Lu2005MiRNAProfiles}. These findings motivate miRNA abundance as a biologically meaningful auxiliary prediction target, although they do not establish that inferred profiles recover every regulatory response.

The practical motivation for inference is the limited availability of matched regulatory measurements in large perturbation-expression resources such as LINCS L1000 \cite{subramanianNextGenCMap2017}. Obtaining matched miRNA profiles requires additional assays, experimental time, and sample material, with costs accumulating across compounds, doses, and time points. Computational inference can reuse existing transcriptomes without repeating these assays for every condition. Recent machine-learning and deep-learning approaches make this strategy increasingly feasible \cite{olgunMiRSCAPE2022,caiSiCmiRAtlas2025}. Here, inference serves as a scalable representation-building strategy; it does not replace direct measurement when testing a specific regulatory mechanism.

Related precedents show how a learned representation can improve prediction when combined with features from the same input. In computer vision, modality hallucination uses paired RGB--depth training data to learn depth-like features from RGB, combining both RGB-derived branches to improve detection \cite{Hoffman2016Hallucination}. In natural language processing, CoVe concatenates word vectors with representations from a supervised machine-translation encoder, improving sentiment analysis and question answering \cite{McCann2017CoVe}. ELMo likewise augments existing token representations to improve question answering and textual entailment, although its pretraining objective is language modelling rather than supervised translation \cite{Peters2018ELMo}. These approaches transfer structure learned from an auxiliary objective without requiring its target modality or labels at deployment. They motivate testing whether miRNA-supervised representations can make transcriptomic structure more useful for downstream prediction. The unresolved question is whether this biologically informed recoding improves pathway classification and MoA retrieval beyond gene expression, generic embeddings, or inferred TF activity.

Although the biological rationale is compelling, accurate computational inference of these regulatory layers remains challenging. For TF activity estimation, a range of methods has been developed, including enrichment-based approaches such as VIPER \cite{alvarezVIPER2016}, matrix factorization methods such as TIGER \cite{chenPadiTIGER2024}, and linear model-based strategies such as Priori \cite{yasharPriori2024}. In parallel, methods for inferring miRNA abundance from transcriptomic data have evolved from early correlation-based models to more sophisticated frameworks such as miRSCAPE \cite{olgunMiRSCAPE2022}, miTEA-HiRes \cite{miTEAHiRes2025}, and our previously developed SiCmiR \cite{caiSiCmiRAtlas2025}, which uses deep learning to capture non-linear mRNA--miRNA dependencies. However, these methods have rarely been optimized or benchmarked side by side in pharmacological settings, where perturbational transcriptomes are often subtle, noisy, and highly context-dependent compared with conventional tissue phenotyping tasks \cite{subramanianNextGenCMap2017,iorioDiscoveryAnticancer2017}. More importantly, the downstream value of inferred regulatory features for drug mechanistic modeling remains unclear.

In particular, it is not well understood how alternative regulatory feature layers compare as transferable inputs for distinct mechanistic tasks, or whether such features can mitigate overfitting and improve robustness.

Here, we present a computational framework for systematically comparing inferred regulatory feature layers in perturbational drug modeling (Figure \ref{fig:workflow}). We optimize HubmiRNet to infer 414 hub miRNAs from L1000 landmark genes and benchmark TF activity inference methods to construct the corresponding regulatory representations. We then evaluate these regulatory representations side by side across two downstream tasks, causal pathway reconstruction and similarity-based MoA retrieval, to determine which layer provides the most robust and transferable benefit beyond gene expression alone. Across these analyses, inferred miRNA representations deliver the most consistent performance gains, whereas TF activities offer more limited incremental value. Together, these results support inferred regulatory layers as a scalable strategy for enhancing the robustness and interpretability of pharmacological modeling, and highlight explicit modeling of the miRNA layer as a particularly promising direction for AI-driven drug discovery.

\section{Results}
\subsection{HubmiRNet Accurately Predicts Hub-miRNA Expression}

To evaluate the downstream contribution of an inferred miRNA-expression layer in drug mechanistic modeling, we first sought to establish a reliable strategy for predicting miRNA abundance from transcriptomic profiles. In our previous work, SiCmiR, we demonstrated the feasibility of reconstructing miRNA expression from matched RNA-seq and miRNA-seq data collected across TCGA and GEO cohorts. Through pan-cancer co-expression analysis, we identified a core set of 414 hub-miRNAs (HubmiRs) that display strong regulatory connectivity and reproducible associations with gene expression modules across diverse biological contexts (Figure \ref{fig:hubmir}a) \cite{caiSiCmiRAtlas2025}. These HubmiRs provide a compact, biologically informed target space for constructing a predicted miRNA-expression representation from L1000 transcriptomic inputs.

We therefore formulated HubmiR inference as a multi-target regression task that maps L1000 landmark gene expression to the 414-dimensional HubmiR profile. To maximize the fidelity of this inferred representation for subsequent mechanistic analyses, we systematically benchmarked multiple neural architectures, including Multi-Layer Perceptrons (MLP), Kolmogorov--Arnold Networks (KAN), Transformers, and DenseNets, against a Residual Network (ResNet)-based design (Figure \ref{fig:hubmir}a). This benchmarking was intended not merely to optimize prediction accuracy, but to ensure that downstream comparisons involving inferred miRNA features would be grounded in the most reliable possible representation.

Across architectures, the ResNet-based model provided the strongest reconstruction performance, and we therefore designated it as HubmiRNet. On the 414-HubmiR task, HubmiRNet achieved a mean Pearson correlation coefficient (PCC) of $87.72\% \pm 0.09\%$ (95\% CI, 87.42--87.98\%) and the lowest reconstruction error (RMSE = 0.491), compared with RMSE values of 0.610 for the MLP and 0.605 for the Transformer (Figure \ref{fig:hubmir}b). Principal component analysis further showed that the predicted and observed HubmiR profiles overlapped in their major axes of variation (Figure \ref{fig:hubmir}c).

To assess whether the performance advantage persisted outside the selected 414-HubmiR target space, we retained the HubmiRNet architecture, changed only the output dimension from 414 to 1,298, and retrained the model on the original SiCmiR task. On the same 1,298-miRNA test set, full-output HubmiRNet achieved a mean PCC of $71.21\% \pm 0.14\%$ (95\% interval, 70.63--71.75\%), compared with $67.30\% \pm 0.23\%$ (95\% interval, 66.35--68.18\%) for the published SiCmiR model \cite{caiSiCmiRAtlas2025}. This corresponded to an absolute PCC increase of 3.91 percentage points. HubmiRNet also reduced RMSE from 0.722 to 0.703. The concordant improvement in correlation and reconstruction error demonstrates that HubmiRNet also surpassed SiCmiR on the original full-miRNA prediction task, extending the reconstruction advantage beyond the 414 selected hub miRNAs.

Together, these results establish HubmiRNet as an accurate inference engine for constructing a biologically informed miRNA representation from L1000 expression data. This step provides the foundation for comparing regulatory feature layers in subsequent perturbational drug modeling analyses and for testing whether the inferred miRNA representation provides complementary predictive utility beyond gene expression under the evaluated data and model regimes.

\subsection{Systematic Benchmarking Identifies a Robust Strategy for Transcription Factor Activity Inference}

To evaluate the downstream contribution of the TF layer in drug mechanistic modeling, we next sought to establish a reliable pipeline for transcription factor activity (TFA) inference from transcriptomic data. Here, TFA was inferred from the coordinated expression of downstream target genes because direct TF-activity measurements were unavailable in the evaluated transcriptomic datasets. The inference strategy therefore affects comparisons between TF-augmented and alternative representations. We therefore performed a systematic benchmark of representative TFA methods and prior knowledge networks (PKNs) before constructing the TF feature space used in later analyses (Figure \ref{fig:tfa}a).

We compared three widely used and conceptually distinct TFA algorithms: Priori, which infers activity through weighted rank aggregation based on TF--target regulatory relationships; TIGER, which applies Bayesian matrix factorization with flexible edge-sign modeling; and VIPER, which estimates TF activity using analytic rank-based enrichment analysis. Each method was evaluated in combination with three commonly used PKNs---CollecTRI, DoRothEA, and DoRothEA (Pancancer)---across three gold-standard perturbation datasets (A375, MCF7, and Holland), in which the perturbed TFs are known. Performance was assessed using three complementary metrics: Success, Accuracy, and Median Rank of the perturbed TF (Figure \ref{fig:tfa}b).

Across datasets and metrics, both the choice of inference algorithm and the prior network substantially affected performance. Among all tested configurations, the combination of Priori with DoRothEA showed the most consistent robustness, achieving the best or second-best result in 8 of 9 benchmark scenarios (3 datasets $\times$ 3 metrics). Configurations using DoRothEA (Pancancer) also performed strongly, ranking second in 7 scenarios. At the algorithmic level, the rank-based methods Priori and VIPER generally outperformed the matrix-factorization-based TIGER approach in these perturbational benchmarks, indicating that robust rank-based summarization may be better suited to recovering TF perturbation signals under noisy expression conditions.

Based on these results, we selected the Priori+DoRothEA configuration for constructing inferred TF activities in all subsequent downstream analyses. This benchmark therefore serves two purposes: it identifies a robust strategy for TF feature construction, and it establishes a fair comparator for the systematic TF-versus-miRNA layer comparisons that follow. While this benchmark is based on TF perturbation datasets rather than direct chemical perturbations, it provides a principled and empirically supported basis for selecting a downstream TFA inference strategy.

\subsection{Systematic Comparison of Regulatory Feature Layers Highlights miRNA Gains in Causal Pathway Inference}

We next asked whether inferred regulatory states improve the recovery of upstream signaling perturbations beyond transcriptomic endpoints alone. Using the PROGENy benchmark, we formulated this task as multiclass classification of 11 canonical cancer pathways from drug-induced expression profiles and evaluated four model classes shown in Figure~\ref{fig:path_performance}: random forest (RF), multilayer perceptron (MLP), Kolmogorov--Arnold network (KAN), and residual network (ResNet).

We then compared Gene, Gene+TFA, Gene+HubmiR, and Gene+TFA+HubmiR. Averaged across the four models and seven runs per model, Gene-only accuracy and macro-F1 were 70.17\% and 62.14\%, respectively, whereas Gene+HubmiR achieved 74.66\% and 66.92\%, corresponding to gains of 4.50 and 4.78 percentage points. Gene+HubmiR achieved the highest mean accuracy and macro-F1 for each of the four models. Adding TFA alone produced smaller and more variable changes, and adding TFA on top of Gene+HubmiR did not increase the mean beyond Gene+HubmiR for any of the four learners.

Thus, the advantage of inferred HubmiR features was reproduced across tree-based, conventional neural, KAN, and residual architectures. This model-spanning improvement identifies the inferred miRNA layer as the most consistently useful regulatory augmentation in the pathway task and motivates the subsequent analysis of whether its contribution reflects generic compression or a complementary representation.

\subsection{HubmiR Defines a Distinct and Complementary Regulatory Representation}

To determine whether the pathway-classification gains reflected generic feature compression, we compared HubmiR with three 414-dimensional controls generated from the same 977 landmark genes: principal component analysis (PCA), an autoencoder (AE), and Gaussian random projection (RP) (Figure~\ref{fig:feature_analysis}a). In RF, adding PCA, AE, or RP produced only modest changes relative to Gene, whereas HubmiR alone achieved a higher mean macro-F1 than Gene and all three generic controls. Gene+HubmiR achieved the highest mean performance overall. In ResNet, all three generic controls and HubmiR alone performed below Gene, whereas Gene+HubmiR markedly exceeded every alternative. Thus, only the combined Gene+HubmiR representation consistently achieved the strongest performance in both model classes. Supplementary Figure S1 extends the comparison with held-out accuracy for all four learners (panel a) and macro-F1 for KAN and MLP (panel b), using Gene, the three generic augmentations, HubmiR alone, and Gene+HubmiR. Gene+HubmiR had the highest mean accuracy in KAN, whereas Gene+PCA slightly exceeded Gene+HubmiR in MLP (75.80\% versus 75.46\%). The macro-F1 comparison likewise favoured Gene+HubmiR in KAN, whereas Gene+PCA had a slightly higher mean than Gene+HubmiR in MLP (69.45\% versus 68.24\%). The AE reconstruction curves reached a stable validation optimum (Supplementary Figure S1c).

We next used regularized canonical correlation analysis (CCA) to measure the strongest shared linear modes between Gene and each derived representation. CCA identifies pairs of feature combinations with maximal correlation, so persistently high canonical correlations indicate closely aligned information across multiple modes. The three generic controls showed uniformly high correlations across the first ten modes, with the yellow PCA, AE, and RP curves remaining at approximately 0.87--0.99 (Figure~\ref{fig:feature_analysis}b). HubmiR retained a strong first canonical mode (mean correlation, 0.92), but its subsequent modes were consistently lower, at approximately 0.76--0.86. This contrast shows that HubmiR preserves important gene-derived signals while departing more substantially from the linear structure retained by generic compression.

Centred-kernel alignment (CKA) provided a complementary view of the overall sample geometry. Unlike CCA, which identifies the most highly correlated feature combinations, CKA compares the global pattern of relationships among samples across two representations. Across the complete 20-split analysis, mean linear and RBF CKA values were 0.28 and 0.33 for HubmiR, compared with 0.56--0.69 for PCA, AE, and RP; Figure~\ref{fig:feature_analysis}c shows seven evenly spaced display splits. The consistent separation under both kernels indicates that HubmiR reorganizes the gene-expression space at both linear and nonlinear levels.

We then used multi-output ridge regression to test how accurately a regularized linear mapping from Gene could reconstruct the 414 HubmiR features. The full gene matrix explained 45\% of held-out HubmiR variance, whereas the 977 HubmiRNet input genes explained 43\% (Figure~\ref{fig:feature_analysis}d). Per-miRNA recovery patterns and alternative aggregate $R^2$ summaries were also similar for the full Gene and 977-gene inputs (Supplementary Figure S1d,e). These results show that most linearly recoverable signal was already present in the landmark-gene input, while more than half of HubmiR variation was not captured by either linear model. Together, CCA, CKA, and ridge regression establish representation-level complementarity: HubmiR retains gene-derived biological signal but reorganizes it through a nonlinear mapping learned from independent paired transcriptomic and miRNA reference data, yielding a distinct and only partially linearly recoverable feature space.

Finally, paired comparisons across RF, MLP, KAN, and ResNet showed positive mean macro-F1 changes when HubmiR was added to Gene and when Gene was added to HubmiR alone (Figure~\ref{fig:feature_analysis}e). The gains relative to both component representations establish that the benefit cannot be attributed solely to either the original gene space or the inferred HubmiR space. Instead, HubmiR provides a biologically informed and task-relevant recoding that complements gene expression across the evaluated model classes.

% \clearpage

\subsection{Inferred HubmiR Signatures Rescue Classification Across Diverse Signaling Pathways}

To determine whether the inferred miRNA layer could resolve individual pathway states missed by gene expression alone, we examined strict rescue cases in which the Gene-only ResNet prediction was incorrect but the Gene+HubmiR prediction was correct for the same held-out sample and run. Across the seven paired ResNet runs used in Figure~\ref{fig:path_performance}, 50 of 595 held-out sample--run occurrences (8.4\%) were rescued, whereas 12 (2.0\%) were harmed, corresponding to a 4.17-fold excess of rescues and a net gain of 38 correctly reclassified occurrences (Supplementary Figure S2a). Rescues outnumbered harms in every pathway and were observed across all 11 pathway classes (Supplementary Figure S2b).

Figure~\ref{fig:path_case} examines nine representative strict-rescue perturbations spanning three signaling axes: basic fibroblast growth factor (bFGF), gefitinib, and epidermal growth factor (EGF) for EGFR; PLX4720, the MEK1 inhibitor PD98059, and MAP2K1 knockdown for MAPK; and 1\% O$_2$, melanoma-conditioned-medium hypoxia, and hypoxia exposure for the hypoxia pathway. Thus, the rescue pattern extended across receptor stimulation, kinase inhibition, genetic knockdown, and oxygen deprivation rather than being restricted to one pathway or perturbation class.

Across all three pathways, the pathway-related PROGENy gene-footprint features showed comparatively weak expression intensities in the rescued samples, leaving the Gene-only model with limited pathway-discriminative signal. By contrast, the corresponding pathway-related HubmiRs showed large absolute expression values across the same samples, indicating pronounced regulatory features for pathway classification. This high-magnitude HubmiR signal was evident across the EGFR, MAPK, and hypoxia perturbations and supplied discriminatory features that enabled the combined model to recover the correct pathway labels. Several miRNAs also showed coherent directions within related perturbations, further supporting their pathway relevance, but the principal contrast was their stronger absolute regulation intensity relative to the displayed gene features.

The selected features were also closely aligned with the biology of their corresponding pathways. All eight genes in each panel were drawn from the experimentally derived PROGENy response footprint for that pathway \cite{schubertPROGENy2018}. The highlighted miRNAs have likewise been linked to the relevant signaling contexts: miR-200c, miR-138-5p, and miR-30d to EGFR-inhibitor response or EGFR/FGFR--MAPK signaling \cite{Lin2021miR200cEGFR,Cui2020miR138EGFR,Li2021miR30dMAPK}; miR-7, miR-145, and miR-664a-3p to EGFR--ERK, MAPK1, or MAPK/ERK regulation \cite{Zhou2014miR7EGFRERK,Yang2018miR145MAPK1,He2021miR664MAPK}; and miR-210, miR-101, and miR-129-2-3p to hypoxia-responsive programs \cite{Liu2017miR210Hypoxia,Kim2014miR101Hypoxia,Zhang2019miR129Hypoxia}. This agreement between the displayed regulatory features and established pathway biology supports a biologically coherent basis for the observed rescue patterns.

Together, these representative cases illustrate a broader pattern across rescued samples from diverse pathways: inferred pathway-related HubmiR signals can remain strongly regulated when the pathway-corresponding gene-level readouts are weak. The inferred miRNA layer therefore provides a pathway-oriented representation that complements the transcriptomic input and supports correct classification across diverse rescued pathway samples. A pathway-wide catalogue of 27 selected strict-rescue profiles, together with their perturbation contexts, Gene-only misclassifications, and feature-level patterns, is provided in Supplementary Figures S2c and S3.

\subsection{Inferred miRNA Signatures Provide Transferable Gains in Similarity-Based Drug MoA Retrieval}

\begin{table*}[!t]
\caption{Drug MoA similarity prediction performance ($\mathrm{AUC}_{0.01}$) across four algorithms, four feature-selection strategies, and six cell lines. Bold and underlined values indicate the best and second-best results, respectively, for each algorithm. LM: landmark gene signature; Mi: inferred miRNA signature; TF: inferred transcription factor signature.}
\centering
\setlength{\tabcolsep}{4pt}
\renewcommand{\arraystretch}{1.15}
\begin{tabular}{>{\centering\arraybackslash}m{2.3cm}>{\centering\arraybackslash}m{2.7cm}>{\centering\arraybackslash}m{1.55cm}>{\centering\arraybackslash}m{1.55cm}>{\centering\arraybackslash}m{1.55cm}>{\centering\arraybackslash}m{1.55cm}>{\centering\arraybackslash}m{1.55cm}>{\centering\arraybackslash}m{1.55cm}}
\toprule
\textbf{Algorithm} & \textbf{Feature Space} & \textbf{A375} & \textbf{A549} & \textbf{HA1E} & \textbf{HCC515} & \textbf{MCF7} & \textbf{VCAP} \\
\midrule
\multirow{4}{*}{\textbf{weightedKS}} & \textbf{LM+TF}    & 0.0459 & 0.0182 & 0.0514 & 0.0090 & 0.0546 & 0.0117 \\
                                   & \textbf{LM only}  & \underline{0.0668} & \underline{0.0272} & \textbf{0.0636} & \underline{0.0343} & \underline{0.0990} & 0.0188 \\
                                   & \textbf{LM+Mi}    & \textbf{0.0803} & \textbf{0.0377} & \underline{0.0622} & \textbf{0.0465} & \textbf{0.1037} & \textbf{0.0264} \\
                                   & \textbf{LM+Mi+TF} & 0.0523 & 0.0169 & 0.0509 & 0.0094 & 0.0586 & \underline{0.0226} \\
\addlinespace[4pt]
\cmidrule(l{8pt}r{8pt}){1-8}
\addlinespace[5pt]
\multirow{4}{*}{\textbf{KS}}       & \textbf{LM+TF}    & 0.0394 & 0.0160 & 0.0275 & 0.0029 & 0.0533 & 0.0165 \\
                                   & \textbf{LM only}  & \underline{0.0662} & \underline{0.0201} & \textbf{0.0643} & \underline{0.0197} & \underline{0.0829} & \underline{0.0203} \\
                                   & \textbf{LM+Mi}    & \textbf{0.0664} & \textbf{0.0202} & \underline{0.0574} & \textbf{0.0286} & \textbf{0.0853} & \textbf{0.0249} \\
                                   & \textbf{LM+Mi+TF} & 0.0590 & 0.0117 & 0.0264 & 0.0126 & 0.0596 & 0.0189 \\
\addlinespace[4pt]
\cmidrule(l{8pt}r{8pt}){1-8}
\addlinespace[5pt]
\multirow{4}{*}{\textbf{XCos}}     & \textbf{LM+TF}    & 0.0540 & 0.0163 & \underline{0.0473} & 0.0241 & 0.0536 & 0.0213 \\
                                   & \textbf{LM only}  & \textbf{0.0919} & \textbf{0.1008} & 0.0464 & \underline{0.0544} & \textbf{0.1269} & \textbf{0.0250} \\
                                   & \textbf{LM+Mi}    & \underline{0.0817} & \underline{0.0945} & \textbf{0.0893} & \textbf{0.0615} & \underline{0.1219} & 0.0202 \\
                                   & \textbf{LM+Mi+TF} & 0.0676 & 0.0208 & 0.0431 & 0.0178 & 0.0594 & \underline{0.0223} \\
\addlinespace[4pt]
\cmidrule(l{8pt}r{8pt}){1-8}
\addlinespace[5pt]
\multirow{4}{*}{\textbf{Zhang}}    & \textbf{LM+TF}    & 0.0382 & 0.0179 & 0.0266 & 0.0057 & 0.0506 & 0.0154 \\
                                   & \textbf{LM only}  & \underline{0.0676} & \underline{0.0211} & \textbf{0.0658} & \underline{0.0235} & \underline{0.0891} & \underline{0.0235} \\
                                   & \textbf{LM+Mi}    & \textbf{0.0736} & \textbf{0.0263} & \underline{0.0620} & \textbf{0.0312} & \textbf{0.0917} & \textbf{0.0250} \\
                                   & \textbf{LM+Mi+TF} & 0.0555 & 0.0132 & 0.0293 & 0.0077 & 0.0589 & 0.0205 \\
\bottomrule
\end{tabular}
\label{tab:MoA_result}
\end{table*}

To test whether the value of inferred regulatory layers extends beyond supervised pathway classification or instead reflects task-specific fitting, we next evaluated a mechanistically distinct task: similarity-based retrieval of drug mechanism of action (MoA) from L1000 perturbational profiles. We compared four feature configurations---landmark genes only (LM), LM+TF, LM+Mi, and LM+TF+Mi---and quantified drug similarity using four representative connectivity-scoring methods spanning enrichment-based, cosine-similarity-based, and signed rank-based approaches. Because the benchmark dataset is strongly imbalanced between positive and negative drug pairs, performance was assessed primarily using the early-retrieval metric $\mathrm{AUC}_{0.01}$.

Across six cell lines and four scoring algorithms, inferred miRNA features again showed the most consistent positive contribution to performance (Table~\ref{tab:MoA_result}). In multiple algorithm--cell-line combinations, LM+Mi outperformed the landmark-only baseline and achieved the strongest overall results, particularly for the weightedKS, KS, and Zhang scoring schemes. By contrast, adding inferred TF activity alone produced limited and often inconsistent benefit. Although the relative ordering of LM and LM+Mi varied under XCos, the overall pattern remained clear: the positive effect of inferred miRNA features was not confined to a single similarity metric, but generalized across multiple retrieval paradigms with some method-dependent variability.

The joint feature space LM+TF+Mi did not systematically improve upon LM+Mi and in several settings performed worse. This pattern suggests that inferred TF activities add little incremental signal once landmark transcriptional profiles and inferred miRNA features are already present, and may in some cases introduce redundancy or dilute more informative structure. Among the evaluated similarity functions, cosine-based approaches generally achieved higher absolute $\mathrm{AUC}_{0.01}$ values than enrichment-based methods; however, this difference in absolute scoring performance did not alter the relative feature-level conclusion that miRNA augmentation provides the most reliable benefit across settings.

To summarize these effects across algorithms and cell lines, we fitted a mixed-effects linear regression model with feature configuration as a fixed effect and algorithm and cell line as grouping factors. Aggregating the full benchmark yielded 24 observations for each feature configuration. Consistent with the table-level patterns, inclusion of inferred miRNA features was associated with a positive overall effect on MoA retrieval performance, whereas TF-only augmentation showed a negligible or negative effect. The coefficient for the combined LM+TF+Mi representation was close to zero, indicating no systematic advantage over simpler feature spaces. Together, these results extend the pathway-inference findings to a retrieval-based MoA setting and strengthen the conclusion that inferred miRNA representations provide the most transferable and practically useful regulatory layer across perturbational drug modeling tasks.

\section{Discussion}

Across pathway classification and MoA retrieval, adding inferred miRNA features improved predictive performance relative to gene expression alone under the evaluated settings. This is compatible with both branches depending on the same observed transcriptome. For a frozen HubmiRNet, $M=f_{\theta}(G)$, so $[G,M]$ adds no independent sample-level measurement to $G$. Its potential benefit instead comes from the mapping learned using paired mRNA--miRNA supervision, which supplies a biologically informed inductive bias. A downstream learner with finite training data and constrained capacity can use this mapping without having to relearn it from pathway labels alone. Modality hallucination and CoVe provide precedents for this distinction between additional observations and additional learned structure \cite{Hoffman2016Hallucination,McCann2017CoVe}. Retaining Gene alongside HubmiR also preserves input variation that the compressed representation may omit. The improvement is therefore a practical property of the representation and learning regime, not a guarantee that deterministic augmentation benefits every model.

The choice of auxiliary target is especially relevant because both downstream tasks depend on the cellular response to a perturbation. MiRNAs participate in regulatory circuits that coordinate and buffer gene-expression programs, making them biologically motivated targets for learning context-sensitive representations \cite{Tsang2007NetworkMotifs,EbertSharp2012}. Our hypothesis is that predicting their abundance emphasizes combinations of transcripts associated with these programs, potentially helping distinguish pathway responses or compounds with shared MoA. HubmiRNet learns these associations from paired expression data; it does not explicitly reconstruct feedback circuits or measure miRNA target repression. The weaker benefit of the evaluated TFA features therefore concerns these particular representations and inference pipelines, rather than the biological importance of TF regulation. This distinction also avoids attributing the TF--miRNA difference entirely to biology when their training objectives and feature-construction procedures differ.

The control analyses help delimit this interpretation. Gene+HubmiR exceeded the generic controls in RF and ResNet macro-F1, whereas PCA slightly exceeded it in MLP, showing that biological supervision is not universally superior. CKA, CCA, and ridge recovery characterize changes in representation geometry and linear accessibility; they do not establish independent molecular information. The comparison with HubmiR alone further supports retaining the original Gene features alongside the learned representation. MoA retrieval tests a distinct consequence of recoding: adding HubmiR changes similarity rankings rather than training a pathway classifier. Concordant gains across classification and retrieval thus suggest utility across two decision rules, without establishing recovery of unmeasured miRNA responses. The evidence supports task-relevant representation transfer, while the contributions of denoising, optimization and biological supervision remain to be separated more fully.

Representative rescue cases illustrate how inferred miRNA features can assist classification when the displayed transcriptomic readouts are weak or ambiguous. Across representative EGFR, MAPK, hypoxia, and supplementary cases, Gene-only models failed to recover the expected pathway identity, whereas Gene+HubmiR models restored correct classification. These examples provide a pathway-spanning illustrations for the broader performance gains observed in our benchmark tasks: endpoint transcriptomic signatures do not always provide a sufficiently distinctive footprint of upstream pathway activity. Weak or partially ``silent'' signatures may arise from transient signaling, compensatory feedback, or incomplete propagation of pathway activity to steady-state mRNA abundance, as exemplified by the dense feedback architecture of EGFR/ERBB signaling and the adaptive transcriptional dynamics of hypoxia responses \cite{AvrahamYarden2011,Liu2017miR210Hypoxia,Kim2014miR101Hypoxia}. More broadly, perturbational transcriptomic signatures contain strong context-dependent and confounding components, including viability-associated effects, that can obscure direct mechanism-of-action relationships \cite{SzalaiEtAl2019}. Under such conditions, inferred HubmiR features can provide a pathway-oriented regulatory recoding that remains informative when canonical gene-expression readouts are weak, diffuse, or difficult to interpret.

These observations have broader implications for pharmacological AI. Much current work seeks better performance by scaling datasets, enlarging models, or refining optimization strategies. Our results suggest an additional and underappreciated axis of improvement: enhancing the biological abstraction of the input representation itself. Our framework uses miRNA-supervised representations to organize transcriptomic inputs according to associations learned from paired molecular profiles. This strategy is conceptually aligned with multi-omics approaches that resolve information across molecular layers, and with recent perspectives emphasizing the utility of perturbation-expression profiles for MoA analysis, pathway activity inference, and quantitative drug modeling \cite{Hasin2017,SzalaiVeres2023}. Because inferred regulatory features are also more readily linked to known control circuits than raw gene-level signatures, such representations may additionally improve interpretability and support mechanism-oriented hypothesis generation.

Several limitations should be acknowledged. First, the miRNA layer in this study was inferred rather than directly measured, and thus should be interpreted as a biologically informed representation for post-transcriptional state rather than a direct readout of regulatory activity. Although HubmiRNet showed strong reconstruction performance, inference error and domain shift may affect downstream utility. Frozen-model validation against independently measured perturbational miRNAs is needed to assess recovery of the underlying molecular response. Second, the limited incremental value of inferred TF activities in our analyses should not be interpreted as evidence that TF regulation is biologically unimportant; rather, it suggests that, within the present feature-construction framework and downstream tasks, TF-derived features are more redundant with transcriptomic inputs than inferred miRNA features. Third, the framework was developed and evaluated in perturbational transcriptomic settings that may not fully capture variation across cell states, doses, or temporal regimes. Fourth, we focused specifically on TFs and miRNAs, whereas other regulatory layers---including phospho-signaling, RNA-binding proteins, and chromatin state---may also encode important mechanistic information, as highlighted by broader multi-omics studies \cite{Hasin2017,VogelMarcotte2012}. Molecular-interaction representations offer another direction: PepPI combines sequence and structural features to predict protein--peptide interactions, illustrating a complementary route to drug-target hypothesis generation \cite{CaoPepPI2025}. Finally, the evaluated tasks represent useful but incomplete proxies of drug action, and further work will be required to determine whether the same principles generalize to target deconvolution, resistance modeling, and therapeutic prioritization.

Taken together, our results show that systematic comparison of regulatory feature layers can reveal which biological abstractions transfer across perturbational drug modeling tasks. By benchmarking TF-based representations and developing HubmiRNet for hub-miRNA inference, we identify inferred miRNA representations as the most informative and transferable regulatory layer in this study. More broadly, these findings suggest that improving pharmacological AI will require not only stronger prediction engines, but also biologically informed representations that better reflect how drug perturbations propagate through layered cellular control systems \cite{Hasin2017,Bartel2018}.

\section{Methods}

\subsection{Datasets and preprocessing}

\subsubsection{Training data for HubmiRNet}
To construct the training dataset for miRNA inference, we used the paired mRNA and miRNA expression compendium established in our previous SiCmiR study \cite{caiSiCmiRAtlas2025}, which integrates matched transcriptomic and miRNA profiles from The Cancer Genome Atlas (TCGA) and Gene Expression Omnibus (GEO). For the primary HubmiRNet analyses, the input space was restricted to the 977 landmark genes defined by the LINCS L1000 platform, and the output space was restricted to 414 hub miRNAs (HubmiRs) previously identified by pan-cancer co-expression network analysis in SiCmiR. For the full-miRNA benchmark in Figure \ref{fig:hubmir}b, the same 977 inputs were paired with all 1,298 miRNA outputs used in the original SiCmiR prediction task.

Gene-expression and miRNA-expression values were transformed using log$_2$(CPM + 1), following the preprocessing protocol established in SiCmiR \cite{caiSiCmiRAtlas2025}. Batch effects across TCGA and GEO cohorts were harmonized using the same pipeline as in the original study. The final dataset contained 9,230 paired profiles, comprising 6,462 samples in the original SiCmiR training partition and 2,768 samples in the independent held-out test partition. Model fitting and selection used the training partition, and final reconstruction metrics were evaluated on the 2,768 held-out samples.

\subsubsection{TF perturbation benchmark datasets}
To benchmark transcription factor activity (TFA) inference methods, we used three perturbation datasets with known ground-truth TF labels.

\begin{enumerate}
    \item \textbf{A375} ($n=50$), consisting of shRNA-mediated TF knockdown experiments.
    \item \textbf{MCF7} ($n=87$), consisting of shRNA-mediated TF knockdown experiments.
    \item \textbf{Holland} ($n=124$), consisting of a mixture of TF knockdown and overexpression experiments.
\end{enumerate}

The A375 and MCF7 datasets were obtained from the validation cohort used in the study \textit{Flexible modeling of regulatory networks improves transcription factor activity estimation}, whereas the Holland dataset was obtained from \textit{Predicting transcription factor activity using prior biological information}. For each sample, the ground-truth label was defined as the TF directly perturbed in the corresponding experiment.

\subsubsection{PROGENy pathway benchmark dataset}
For the pathway-classification task, we used the curated PROGENy benchmark dataset \cite{schubertPROGENy2018}, which contains pathway-responsive transcriptomic signatures generated from perturbation experiments targeting 11 canonical signaling pathways: EGFR, MAPK, PI3K, VEGF, JAK--STAT, TGF$\beta$, TNF$\alpha$, NF$\kappa$B, Hypoxia, p53, and Trail.

Following the original PROGENy framework, gene-level features were represented as Z-scores of expression change relative to matched controls. The final classification dataset used in this study contained 565 samples used after preprocessing, each annotated with a single pathway label corresponding to the experimentally perturbed signaling axis. Gene identifiers were mapped to a common namespace. Samples with missing values in required features were imputed by replacing missing TF activity values with 0.

\subsubsection{Drug mechanism-of-action benchmark dataset}
To evaluate similarity-based retrieval of drug mechanism of action (MoA), we integrated compound annotations from the Drug Repurposing Hub with perturbational expression profiles from the LINCS L1000 resource. Compound annotations, including MoA and target labels, were downloaded from the Drug Repurposing Hub (\texttt{https://clue.io/repurposing}; archived on May 16, 2018). After excluding compounds with missing or ambiguous annotations, we retained compounds that could be paired using exact matches of both annotated MoA and target, yielding a curated set of 1,921 compounds.

Perturbational signatures for these compounds were obtained from LINCS Level 5 data (GSE92742). We retained six cell lines with relatively high compound coverage: HCC515, HA1E, MCF7, VCAP, A375, and A549. The resulting benchmark comprised HCC515 (596 drugs; 1,304 profiles), HA1E (632 drugs; 1,529 profiles), MCF7 (718 drugs; 3,711 profiles), VCAP (630 drugs; 1,915 profiles), A375 (637 drugs; 1,287 profiles), and A549 (635 drugs; 2,225 profiles).

For each perturbational signature, we constructed four feature configurations: landmark gene expression only (LM), LM plus inferred TF activity (LM+TF), LM plus inferred miRNA features (LM+Mi), and the combined representation (LM+TF+Mi). Similarity evaluation was performed within each cell line using individual perturbational signatures as the basic retrievable units. Positive pairs were defined as signature pairs whose corresponding compounds shared both the same annotated MoA and the same annotated target, whereas all other eligible pairs were treated as negatives. This benchmark design follows prior large-scale MoA retrieval settings based on perturbational expression similarity, while adapting the feature space to include inferred regulatory layers.

\subsection{Construction of inferred regulatory layers}

\subsubsection{HubmiR selection}
The target space for miRNA inference consisted of 414 HubmiRs previously defined in the SiCmiR framework \cite{caiSiCmiRAtlas2025}. These hub miRNAs were selected from a pan-cancer miRNA--mRNA co-expression network as a compact subset of highly connected and reproducibly co-regulated miRNAs. In the present study, this set was reused without re-selection.

\subsubsection{HubmiRNet architecture}
We developed HubmiRNet, a residual neural network for multi-target regression from the 977-dimensional L1000 landmark-gene input space to the 414-dimensional HubmiR output space. The model consists of an input projection layer from 977 to 4096 dimensions, followed by three residual blocks and a final linear projection to 414 outputs. For the full-miRNA benchmark, the same model family was configured with a final projection to 1,298 outputs.

Each residual block is defined as
\begin{equation}
y = \sigma\left(\mathrm{BN}_2\left(\mathrm{Linear}_2\left(\mathrm{Dropout}\left(\sigma\left(\mathrm{BN}_1\left(\mathrm{Linear}_1(x)\right)\right)\right)\right)\right) + x\right),
\end{equation}
where $x$ is the block input, $\sigma$ denotes the ReLU activation function, $\mathrm{BN}$ denotes one-dimensional batch normalization, and dropout was applied with probability $p = 0.35$.

\subsubsection{HubmiRNet training and benchmark models}
HubmiRNet was trained by minimizing mean squared error (MSE) between predicted and observed HubmiR expression values. Training used stochastic gradient descent (SGD) with learning rate 0.4, weight decay $2 \times 10^{-4}$, and momentum 0. The learning rate followed a cosine annealing schedule over 2,500 epochs. Weights were initialized from a normal distribution with $\mu=0$ and $\sigma=0.01$. The batch size was 64. The final model checkpoint for each run was selected according to the lowest validation loss.

To assess training stability, the complete HubmiRNet training procedure was repeated seven times using different random seeds. Reported model performance corresponds to mean $\pm$ s.d. across the seven independent runs.

We benchmarked HubmiRNet against four alternatives: MLP, Transformer, KAN and DenseNet. All benchmark models used the same training, validation and test partitions and were trained under comparable optimization settings where applicable. Key architectural settings for all benchmark models are provided in the Supplementary Methods.

\subsubsection{Full-miRNA benchmark against SiCmiR}
To determine whether the HubmiRNet performance advantage extended beyond the selected 414 HubmiRs, we retained the same architecture and changed only the final output dimension from 414 to 1,298. The model was retrained using the original SiCmiR training and test partitions, preprocessing, 977 input genes, and 1,298 miRNA prediction targets \cite{caiSiCmiRAtlas2025}. SiCmiR and full-output HubmiRNet were evaluated on the same test samples and the same ordered miRNA set. Pearson correlation coefficient (PCC) was calculated separately for each miRNA across test samples and then summarized across the 1,298 outputs. Root mean square error (RMSE) was calculated between predicted and observed miRNA-expression values on the same held-out test set. Higher PCC and lower RMSE indicate greater reconstruction accuracy.

\subsubsection{Benchmarking of TF activity inference methods}
To construct the TF regulatory layer, we benchmarked three representative TFA inference methods: VIPER, TIGER, and Priori. VIPER was run using the standard R package implementation with default parameters; TIGER was run using the original author implementation; and Priori was run using the implementation provided by its authors.

Each method was evaluated in combination with three prior knowledge networks (PKNs): CollecTRI, DoRothEA, and DoRothEA (Pancancer). For DoRothEA-based analyses, we retained confidence levels C and above. We did not filter TFs by regulon size, in order to assess robustness across heterogeneous target-set sizes.

Benchmarking was performed on the A375, MCF7, and Holland perturbation datasets described above. For each sample, inferred TF activities were ranked and compared against the known perturbed TF.

\subsubsection{Evaluation metrics for TF activity inference}
We evaluated TFA inference using three metrics: Success, directional accuracy, and median rank.

A perturbation was counted as a \textbf{Success} under the ``top 30\%'' criterion if:
\begin{itemize}
    \item for TF overexpression, the perturbed TF ranked in the top 30\% of TFs within the sample, or the sample ranked in the top 30\% of samples for that TF; or
    \item for TF knockdown, the perturbed TF ranked in the bottom 30\% of TFs within the sample, or the sample ranked in the bottom 30\% of samples for that TF.
\end{itemize}

Directional accuracy was defined as whether the sign of inferred TF activity agreed with the perturbation direction (positive for overexpression and negative for knockdown), relative to untreated control samples. Median rank was computed as the median within-sample rank of the perturbed TF across all samples. The ``Accuracy'' metric reported in Results corresponds to success rate.

Based on this benchmark, Priori + DoRothEA was selected as the default pipeline for constructing inferred TF features in all downstream analyses.

\subsection{Downstream task 1: pathway classification}

\subsubsection{Feature construction}
For each sample in the PROGENy dataset, we constructed four feature matrices:
\begin{enumerate}
    \item \textbf{Genes (G):} gene-expression Z-scores for 29,045 genes;
    \item \textbf{Genes + TFA (G+T):} concatenation of gene-expression features and inferred TF activities;
    \item \textbf{Genes + HubmiR (G+M):} concatenation of gene-expression features and inferred HubmiR features;
    \item \textbf{Genes + TFA + HubmiR (G+T+M):} concatenation of all three feature layers.
\end{enumerate}

Feature normalization was performed separately for each training split. Specifically, Z-score normalization parameters were estimated using the training set only and then applied to the corresponding validation and test sets to avoid information leakage.

\subsubsection{Classification models}
We trained four classifiers to predict the perturbed pathway identity across 11 classes:
\begin{enumerate}
    \item \textbf{Random Forest (RF):} 300 trees, square-root feature sampling, and no maximum depth constraint;
    \item \textbf{Multilayer Perceptron (MLP):} a fully connected input layer with 256 hidden units and ReLU activation, followed by an 11-class linear output layer;
    \item \textbf{Kolmogorov--Arnold Network (KAN):} a two-layer KAN classifier with 128 hidden units and an 11-class output layer;
    \item \textbf{ResNet classifier:} an input projection to 256 hidden units, batch normalization, three fully connected residual blocks with dropout 0.3, and an 11-class linear output layer.
\end{enumerate}

The neural classifiers minimized cross-entropy loss using AdamW. MLP and KAN used a learning rate of $10^{-3}$, batch size 64, a maximum of 60 epochs, and early-stopping patience of 12 epochs. ResNet used a learning rate of $10^{-4}$, weight decay $10^{-3}$, batch size 64, a maximum of 160 epochs, and patience of 24 epochs. For every neural run, the checkpoint with the highest validation macro-F1 was reloaded before held-out test evaluation. Implementations used PyTorch v2.3.0 and Scikit-Learn v1.4.2.

\subsubsection{Data splitting, repeated training, and evaluation}
The pathway benchmark dataset was split into 70\% training, 15\% validation, and 15\% test sets using stratification by pathway label. Each runs used different random seeds, with identical samples across feature configurations within each run.

\subsection{Feature-space control and complementarity analyses}

\subsubsection{Matched embedding controls and downstream evaluation}
To distinguish the HubmiR representation from generic feature compression, we constructed three 414-dimensional controls from the same 977 landmark-gene inputs used by HubmiRNet. The controls were generated from the same frozen 9,230-profile mRNA reference cohort used by SiCmiR, preserving its original partition of 6,462 training and 2,768 held-out samples \cite{caiSiCmiRAtlas2025}. No PROGENy samples were used to fit these transformations. Principal-component analysis (PCA) used randomized singular-value decomposition with seven power iterations and was fitted on the 6,462 reference-training profiles. The autoencoder (AE) comprised a 977--1,024--414 encoder and a symmetric decoder with ReLU activations \cite{HintonSalakhutdinov2006AE}. It was trained on the 6,462 reference-training profiles to reconstruct standardized inputs using AdamW (learning rate $10^{-3}$, weight decay $10^{-5}$, batch size 128), a maximum of 100 epochs, and validation patience of 12 epochs; checkpoint selection used reconstruction loss on the 2,768 held-out reference profiles. Gaussian random projection (RP) provided a structure-agnostic control; five 414-dimensional projections generated with fixed seeds 101, 202, 303, 404, and 505 were evaluated and averaged within each run.

We compared Gene, Gene+PCA, Gene+AE, Gene+RP, HubmiR alone, and Gene+HubmiR using the same stratified partitions of the 565 PROGENy samples. Each split contained 395 training, 85 validation, and 85 test samples. Corresponding feature-space comparisons therefore used identical samples in every partition. RF used 300 trees with square-root feature subsampling. The MLP used one hidden layer with 256 units, whereas the KAN used two layers with 128 hidden units \cite{liuKAN2024}. Both models were trained with AdamW (learning rate $10^{-3}$, batch size 64, maximum 60 epochs, and validation patience 12). The ResNet used a 256-dimensional projection, three residual blocks, dropout of 0.3, and AdamW (learning rate $10^{-4}$, weight decay $10^{-3}$, batch size 64, maximum 160 epochs, and validation patience 24). For MLP, KAN, and ResNet, the parameter state with the highest validation macro-F1 was saved during training. This saved state, rather than the state from the final training epoch, was used to predict the 85 test samples.

Embedding-control performance was assessed by test-set macro-F1 with RF and ResNet across seven existing display runs per configuration. To test whether the contribution generalized across learners, paired Gene+HubmiR minus Gene and Gene+HubmiR minus HubmiR-only differences were evaluated for RF, MLP, KAN, and ResNet. Each contrast comprised the same seven paired runs, with identical test samples used for the two feature spaces within each pair. Downstream model hyperparameters were not re-tuned for individual feature representations. Supplementary Figure S1a reports held-out accuracy for RF, MLP, KAN, and ResNet, and panel b reports macro-F1 for KAN and MLP. Gene and Gene+HubmiR metrics were reused directly from Figure~\ref{fig:path_performance}, and HubmiR-only metrics were reused from the existing utility analyses. KAN and MLP use the same seven frozen splits across all six configurations; their generic augmentations use unchanged model hyperparameters. RF and ResNet accuracy uses the same configuration-specific display runs as the macro-F1 distributions in Figure~\ref{fig:feature_analysis}a. RP metrics were averaged over five projection seeds within each run. The AE training and validation reconstruction losses were retained for all 66 completed epochs; the minimum validation MSE occurred at epoch 54 (Supplementary Figure S1c).

\subsubsection{Representation geometry and linear recoverability}
We quantified shared representation geometry using centred-kernel alignment (CKA) \cite{Kornblith2019CKA} and regularized canonical correlation analysis (CCA) \cite{Hotelling1936CCA}. These analyses used the same 20 frozen stratified splits of the 565 PROGENy samples. In each split, the training partition comprised 395 samples and the held-out test partition comprised 85 samples; the separate 85-sample validation partition was not used for CKA or CCA. Gene and each target representation were standardized using means and standard deviations estimated from the training partition. For CCA, each representation was reduced independently to 64 principal components fitted on the training samples. Ten regularized canonical modes were then learned from the training samples with a ridge parameter of 0.1. The learned canonical directions were applied to the test samples, and the correlation for each mode was calculated exclusively on those 85 samples. Mean test-set correlations and bootstrap 95\% confidence intervals were calculated across all 20 splits. For CKA, linear and radial-basis-function (RBF) kernel matrices were calculated from the standardized test samples. RBF bandwidths were estimated from the corresponding training representation. Figure~\ref{fig:feature_analysis}c shows seven evenly spaced display splits from the complete 20-split CKA analysis.

To quantify linear recoverability, we fitted multi-output ridge regression \cite{HoerlKennard1970Ridge} to predict all 414 HubmiRs from either the full 29,045-gene matrix or the 977-gene HubmiRNet input. Ridge regression used the same 20 PROGENy splits described above. Within each split, the penalty was selected by fivefold cross-validation using only the 395 training samples. The selected model was then refitted on all 395 training samples and evaluated on the 85 test samples. The separate validation partition was not used for model fitting or penalty selection in this analysis. Test-set predictability was summarized as variance-weighted $R^2$ across the 414 outputs. Figure~\ref{fig:feature_analysis}d shows the same seven evenly spaced display splits used for CKA; results from all 20 splits were retained in the complete analysis. For the supplementary analysis, held-out $R^2$ was additionally summarized for each miRNA as the median across the 20 splits and at the aggregate level as variance-weighted, uniform-average, and median per-miRNA $R^2$. Aggregate values are reported as the mean $\pm$ s.d. across splits (Supplementary Figure S1d,e). CKA, CCA, and ridge predictability were interpreted as complementary measures of shared geometry and linear recoverability.

\subsection{Rescued-case analysis for pathway inference}

To investigate cases in which inferred miRNA features corrected pathway predictions missed by transcriptome-only models, we analyzed the frozen held-out predictions from the paired Gene-only and Gene+HubmiR ResNet classifiers. For each sample--run occurrence, the two predictions were assigned to one of four mutually exclusive states: both correct, both incorrect, \textbf{rescued} (Gene-only incorrect and Gene+HubmiR correct), or \textbf{harmed} (Gene-only correct and Gene+HubmiR incorrect). Overall and pathway-specific counts were aggregated across the 595 paired held-out occurrences from the seven ResNet runs used in Figure~\ref{fig:path_performance} (Supplementary Figure S2a,b).

For the focused analysis in Figure~\ref{fig:path_case}, we selected three representative strict-rescue samples for each of EGFR, MAPK, and hypoxia (nine samples in total). The displayed sample identifiers were EGFR.E-GEOD-14256.1, EGFR.E-GEOD-20854.4, and EGFR.E-GEOD-33442.1; MAPK.E-GEOD-17089.4, MAPK.E-GEOD-18232.1, and MAPK.E-GEOD-53091.1; and Hypoxia.E-GEOD-28603.1, Hypoxia.E-GEOD-33115.2, and Hypoxia.E-GEOD-65168.1. To construct the broader supplementary atlas, these nine profiles were retained and up to three distinct strict-rescue profiles were selected for each remaining pathway. For each additional pathway, profiles from the run containing the largest number of distinct rescues were prioritized, after which any remaining positions were filled according to rescue-minus-harm stability across runs. Pathways with fewer than three distinct rescues were not padded with non-rescue profiles. This procedure yielded 27 selected profiles spanning all 11 pathways (Supplementary Figures S2c and S3).

For each pathway, eight pathway-associated genes were drawn from the top 60 official PROGENy response-footprint genes shared across the three displayed samples \cite{schubertPROGENy2018}. The selected genes were PCMTD2, ELF2, PNP, NBR1, POT1, SYF2, BSDC1, and KIAA0430 for EGFR; PNP, ETV4, TCOF1, CCND1, PBLD, EPHA2, ECH1, and PHLDA2 for MAPK; and GPI, VKORC1, PLOD1, CA9, TMEM45A, PDK3, ALDOC, and ALDOA for hypoxia. Three inferred HubmiRs were prioritized for each pathway according to their documented pathway relevance and their pronounced absolute regulation intensities across the corresponding rescued samples: miR-200c-3p, miR-138-5p, and miR-30d-3p for EGFR; miR-7-5p, miR-145-5p, and miR-664a-3p for MAPK; and miR-210-3p, miR-101-3p, and miR-129-2-3p for hypoxia. Feature selection was performed only for visualization and did not enter classifier training or performance evaluation.

For the heatmaps in Figure~\ref{fig:path_case} and Supplementary Figure S3, each selected feature value was expressed as a training-fold empirical-percentile deviation, \mbox{$d=2\{F_{\mathrm{train}}(x)-0.5\}$}, where $F_{\mathrm{train}}$ is the feature-specific empirical cumulative distribution estimated exclusively from the corresponding outer-training fold. The scale ranges from $-1$ to $+1$, with zero denoting the training-fold median. When a profile was rescued in more than one run, its displayed value was averaged across the corresponding strict-rescue occurrences. This transformation was used only for visualization and did not alter classifier inputs or predictions.

\subsection{Downstream task 2: similarity-based MoA retrieval}

\subsubsection{Feature spaces for similarity analysis}
For each drug perturbation profile, we constructed four feature spaces: LM, LM+TF, LM+Mi, and LM+TF+Mi, where LM denotes the 977 landmark-gene signature, TF denotes the inferred TF activity vector, and Mi denotes the inferred HubmiR vector. Feature values were obtained from log-transformed expression values followed by per-feature Z-score normalization. For combined feature spaces, concatenated vectors were jointly ranked for downstream similarity scoring.

\subsubsection{Similarity scoring methods}
We evaluated four connectivity scoring methods: KS, weightedKS, XCos, and ZhangScore. Let $\mathbf{x} = (x_1,\dots,x_n)$ denote a query signature and $\mathbf{y} = (y_1,\dots,y_n)$ denote a reference signature across $n$ ranked features. Following the Connectivity Map framework, the query signature was decomposed into up-regulated and down-regulated feature sets, denoted by $U(\mathbf{x})$ and $D(\mathbf{x})$, respectively. These methods quantify concordance between the ranked reference signature and the query signature using complementary principles, including Kolmogorov--Smirnov enrichment, weighted enrichment, cosine similarity, and signed rank-based association. Detailed mathematical definitions of all scoring functions are provided in Supplementary Methods Section~S2.

\subsubsection{MoA retrieval evaluation}
For each query perturbation profile, all eligible reference profiles within the same cell line were ranked by similarity score. Self-matches were removed prior to evaluation. Ranked lists were then evaluated against the positive/negative pair labels defined above.

Because positive drug pairs were sparse relative to negatives, we used the early-retrieval metric $\mathrm{AUC}_{0.01}$ as the primary evaluation statistic. For each combination of cell line, scoring algorithm, and feature configuration, retrieval performance was computed deterministically from the full ranked list of eligible reference profiles. Reported values therefore correspond to single benchmark estimates rather than averages over repeated runs.

\subsubsection{Mixed-effects statistical analysis}
To summarize retrieval performance across cell lines and scoring methods, we fitted a linear mixed-effects model with $\mathrm{AUC}_{0.01}$ as the response variable. The fixed effects were two binary indicators, \texttt{use\_miRNA} and \texttt{use\_TF}, together with their interaction term, so that the model explicitly tested the marginal contribution of adding inferred miRNA features, inferred TF features, and their joint inclusion. The model formula was
\[
\mathrm{AUC}_{0.01} \sim \mathrm{use\_miRNA} * \mathrm{use\_TF}.
\]
Model fitting was performed using the Python package \texttt{statsmodels} (\texttt{MixedLM}), with scoring algorithm specified as the grouping factor and cell line included as an additional variance component. Statistical significance was assessed using Wald tests. The full benchmark comprised 96 observations in total, corresponding to 6 cell lines $\times$ 4 algorithms $\times$ 4 feature configurations.

\subsection{Implementation and reproducibility}

All deep-learning models were implemented in PyTorch v2.3.0. Classical machine-learning analyses were implemented using Scikit-Learn v1.4.2, and TFA benchmarking was performed using the corresponding original software packages for VIPER, TIGER, and Priori. All analyses were run on a computational cluster equipped with NVIDIA RTX 3090 GPUs.

To ensure reproducibility, random seeds were fixed for data splitting, model initialization, and training. Analysis-specific repeat counts are reported in the corresponding Methods subsections and figure legends. The complete source code, including preprocessing scripts, model definitions, and evaluation pipelines, is available at \url{https://github.com/XinCao02/MIRCID}.

We also developed a publicly accessible MIRCID web interface at \url{https://awi.cuhk.edu.cn/~MIRCID/}. Users can upload normalized gene-expression matrices, infer miRNA representations and TF target scores, and compare pathway predictions and drug MoA retrieval across feature configurations. The interface provides example inputs and downloadable CSV results for exploratory analysis. Deployment-specific models, TF scoring and compound-reference coverage are documented in the interface and should be distinguished from the benchmark protocols evaluated in this study.

\section*{Data availability}
The datasets supporting this study are available at \url{https://zenodo.org/records/22544755}.

\section*{Code availability}
The custom code used for this study is available at \url{https://github.com/XinCao02/MIRCID}. The MIRCID web interface is available at \url{https://awi.cuhk.edu.cn/~MIRCID/}.

\section*{Additional information}
\textbf{Competing interests}: the authors declare no competing interests.

\section*{Acknowledgements}

We would like to express our sincere gratitude to the Vincent \& Lily Woo Foundation for their generous support of the Vincent \& Lily Woo Fellowship in Memory of Dr Albert Wong. This fellowship, endowed by the Vincent \& Lily Woo Foundation, is provided through MCMIA Foundation Limited, and we are deeply grateful for their contribution to our research.

\section*{Author contributions statement}

X.Cao co-designed the study, implemented the computational pipeline for miRNA inference and pathway analyses, and co-wrote the manuscript. Y.C. co-designed the study, led the overall execution, and co-wrote the manuscript. J.X. designed the TFA framework, led TFA analyses, and contributed to other study components. Z.Z. designed and implemented the MoA analyses. X.Cheng, S.W. and Y.Z. contributed to TFA analyses. X.Cai contributed to the design and implementation of miRNA inference. S.C. contributed to miRNA inference analyses. Z.Zhu contributed to pathway inference analyses. X.J. contributed to TFA analyses. H.-Y.H. and Y.-C.-D.L. provided scientific guidance and project supervision. T.Y. and H.-D.H. supervised the study, provided overall direction, and H.-D.H. served as the corresponding authors. All authors discussed the results and approved the final manuscript.

\section*{Biographical Note}

Xin Cao and Ziyue Zhang are Master of Philosophy students in Prof. Hsien-Da Huang's laboratory at The Chinese University of Hong Kong, Shenzhen.
Yigang Chen, Jiatong Xu and Xiang Ji received their PhD degrees from The Chinese University of Hong Kong, Shenzhen, under the supervision of Prof. Hsien-Da Huang.
Xiang Ji, Zihao Zhu and Shidong Cui, are PhD students in Prof. Hsien-Da Huang's laboratory at The Chinese University of Hong Kong, Shenzhen.
Shenyu Wang, Xiang Cheng, and Yangyi Zhang are undergraduate students at The Chinese University of Hong Kong, Shenzhen.
Hsi-Yuan Huang is a research fellow at the School of Medicine and the Warshel Institute of Computational Biology, The Chinese University of Hong Kong, Shenzhen. His research interests include multi-omics data analysis, gene regulation, regulatory RNAs and machine learning.
Yang-Chi-Dung Lin is a research fellow at the School of Medicine and the Warshel Institute of Computational Biology, The Chinese University of Hong Kong, Shenzhen. His research interests include multi-omics data analysis, gene regulation, regulatory RNAs and machine learning.
Tianwei Yu is a professor and associate dean at the School of Data Science, at The Chinese University of Hong Kong, Shenzhen. His research group majorly focuses on bioinformatics, statistics \& machine learning, with applications in metabolomics, pharmacogenomics and systems biology.
Hsien-Da Huang is a presidential chair professor and associate dean at the School of Medicine, and the executive director of Warshel Institute for Computational Biology, The Chinese University of Hong Kong, Shenzhen. His research group majorly focuses on biological multi-disciplinary research topics, including Bioinformatics, Genomics, Metagenomics \& Microbiome, Intelligent Biomedical Technologies (Drug Development, Genetic Test, \& Precision Medicine), AI \& Machine Learning and Biological Database Design \& Development.

\section*{Key Points}

\begin{itemize}
    \item Inferred regulatory feature layers improve perturbational drug modeling beyond endpoint gene expression alone.
    \item HubmiRNet enables robust inference of hub-miRNA representations from L1000 transcriptomic profiles.
    \item Inferred miRNA representations provide more consistent and transferable gains than inferred transcription factor activities.
    \item Illustrative rescue cases highlight the utility of inferred miRNA features when transcriptomic signatures are weak or ambiguous.
    \item Systematic comparison of regulatory abstractions provides a practical framework for representation selection in mechanistic drug modeling.
\end{itemize}

\section*{Funding}

This work was supported by the Shenzhen Medical Research Fund [B2602059]; the Science and Technology Plan Project of Shenzhen [ZDCY20250901102000001]; the Shenzhen Fundamental Research Program [JCYJ20250604141235046; JCYJ20250604141041017 and JCYJ20250604183741054]; the Guangdong S\&T Programme [2024A0505050001; 2024A0505050002]; and Warshel Institute for Computational Biology funding from Shenzhen City and Longgang District [LGKCSDPT2025001]; the Better Way Group - Chinese University of Hong Kong (Shenzhen) Warshel Joint Laboratory for Skin Health and Active Molecule Innovation [2024E0087].

\bibliography{main}

% Insert supp
\beginsupplement

\section*{S1. Supplementary methods for Machine Learning experiments}

\subsection*{S1.1 HubmiR inference data split and preprocessing details}

HubmiR inference was formulated as a multi-target regression problem from transcriptomic input features to 414 HubmiR outputs. Data preprocessing steps not explicitly repeated here follow the released dataset resources and preprocessing scripts in the public repository and data archive.

The full HubmiR inference dataset contained 9,230 paired samples, with 6,462 training samples and 2,768 test samples. Normalization and transform parameters were fit on the training split only and then applied to held-out data.

\subsection*{S1.2 HubmiR benchmark protocol and model-selection rules}

We benchmarked MLP, ResNetMLP, TransformerMLP, KAN, and DenseNet model families under a two-stage model-development workflow: (i) random search across a broad hyperparameter space and (ii) manual tuning around promising regions.

The searched hyperparameter space included hidden size, number of blocks/layers, dropout rate, optimizer type, learning rate, weight decay, learning-rate decay strategy (including milestone-based schedules and cosine annealing), cosine-annealing hyperparameters, Transformer attention-head number and embedding dimension, and KAN-specific grid size and spline order.

Architecture and hyperparameter selection was based on validation PCC; ties were resolved by lower validation loss. Within each final training run, checkpoint selection followed the main-text rule of minimum validation mean squared error.

\subsection*{S1.3 HubmiR model architectures}

\textbf{ResNetMLP (HubmiRNet).}
The HubmiRNet architecture consists of an input projection layer (\texttt{Linear(input\_size, hidden\_size)}), followed by \texttt{BatchNorm1d}, \texttt{ReLU}, and dropout. The core network contains residual fully connected blocks, each structured as:
\[
\begin{aligned}
\mathrm{Linear} \rightarrow \mathrm{BatchNorm1d} \rightarrow \mathrm{ReLU} \rightarrow \mathrm{Dropout} \rightarrow {}\\
\mathrm{Linear} \rightarrow \mathrm{BatchNorm1d} \rightarrow \mathrm{Skip\ connection} \rightarrow \mathrm{ReLU}.
\end{aligned}
\]
The output layer is a final linear projection to the 414-dimensional HubmiR target space.

\textbf{MLP baseline.}
The MLP baseline is a three-hidden-layer fully connected network with repeated blocks of \texttt{Linear--ReLU--BatchNorm1d--Dropout}, followed by a linear output layer.

\textbf{KAN baseline.}
The KAN model uses stacked KAN layers over dimensions
\[
[\mathrm{input\_size},\ \mathrm{hidden\_size},\ \mathrm{hidden\_size},\ \mathrm{output\_size}],
\]
with tunable spline-grid hyperparameters (including grid size and spline order).

\textbf{TransformerMLP baseline.}
The Transformer-style baseline first projects input features into a hidden embedding, then applies repeated self-attention blocks with residual and feed-forward sublayers, and finally maps the representation to output targets via a linear projection.

\textbf{DenseNet baseline.}
The DenseNet baseline uses stacked fully connected blocks (\texttt{Linear--BatchNorm1d--ReLU--Dropout}) with matched depth and width scale for fair comparison, followed by a linear output layer.

\begin{figure*}[p]
    \centering
    \includegraphics[width=0.96\textwidth]{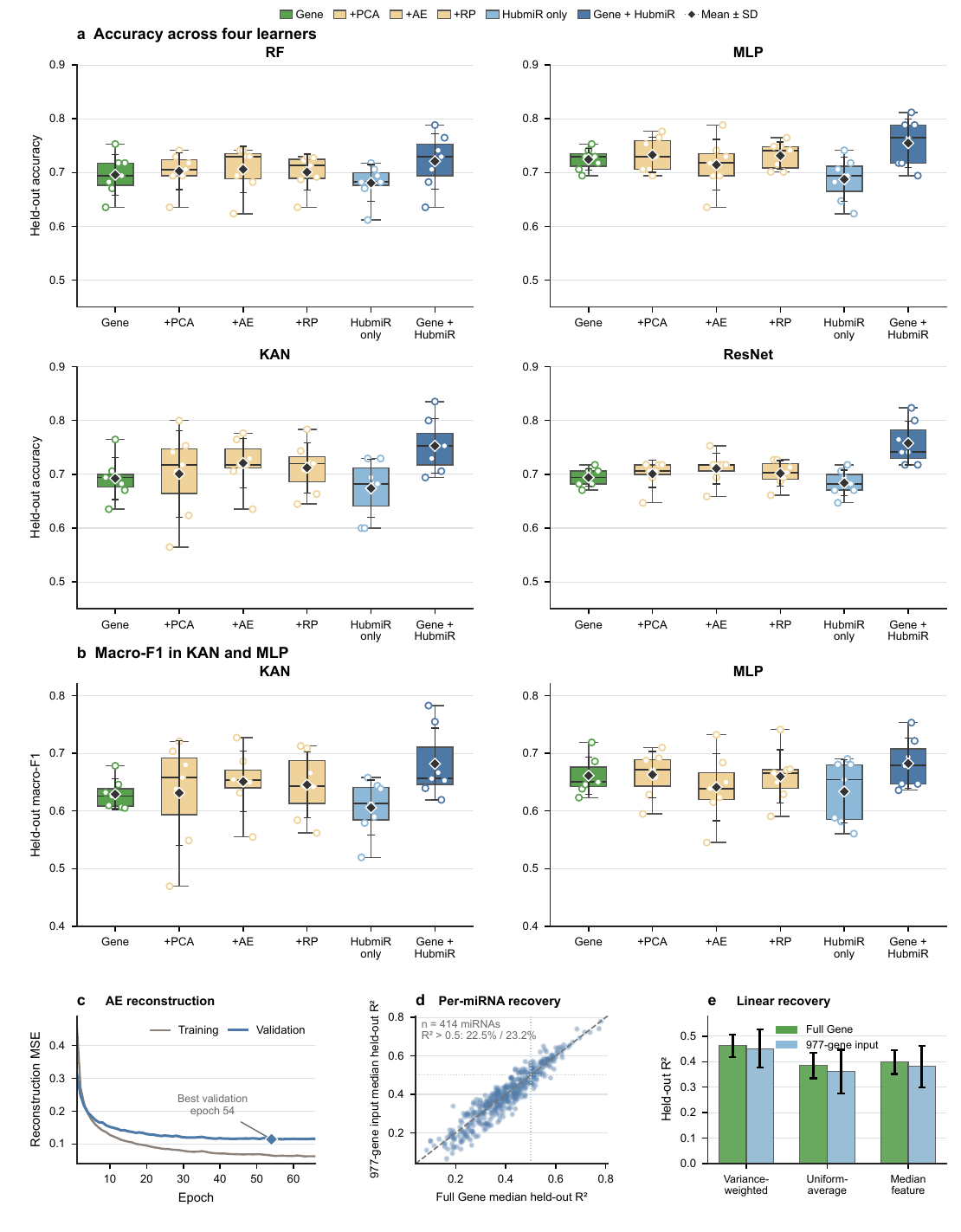}
    \caption[Embedding-control performance across learners and linear recoverability of inferred HubmiR features.]{\textbf{Embedding-control performance across learners and linear recoverability of inferred HubmiR features.} \textbf{a,} Held-out pathway-classification accuracy for RF, MLP, KAN, and ResNet. \textbf{b,} Held-out macro-F1 for KAN and MLP. Six configurations are compared: Gene, Gene augmented with PCA, AE, or RP, HubmiR alone, and Gene+HubmiR ($n=7$ runs per configuration). Gene and Gene+HubmiR reuse Figure 4 results. RP denotes the mean across five projection seeds within each run. Boxes show medians and interquartile ranges, whiskers the observed range, circles individual runs, and diamonds the mean $\pm$ s.d. \textbf{c,} AE training and validation reconstruction MSE, with the minimum validation error marked. \textbf{d,} Median held-out $R^2$ across 20 splits for 414 inferred miRNAs predicted from full Gene and the 977-gene input; the dashed line denotes equal recovery. \textbf{e,} Aggregate held-out $R^2$ across 20 frozen splits for cross-validated multi-output ridge models; bars show the mean $\pm$ s.d.}
    \label{fig:supp_feature_analysis}
\end{figure*}

\subsection*{S1.4 HubmiR final training configuration}

For the selected HubmiRNet setting, training used stochastic gradient descent with learning rate $0.4$, weight decay $2\times10^{-4}$, batch size $64$, dropout $0.35$, and cosine annealing learning-rate scheduling (\texttt{T\_max} set to 2,500 total epochs and \texttt{eta\_min}=0). The input was projected to a 4,096-dimensional hidden space, followed by three residual blocks; each block contained two 4,096-dimensional fully connected layers.

The training objective was mean squared error. The best checkpoint was selected by validation performance according to the main-text checkpoint rule. The complete training procedure was repeated seven times using different random seeds, and performance was summarized as mean $\pm$ s.d. across the seven independent runs.

\subsection*{S1.5 TFA benchmarking details}

Three TFA inference methods were benchmarked: VIPER, TIGER, and Priori. For DoRothEA-based analyses, only interactions with confidence level C or above were retained (DoRothEA $\ge$ C).

We used three evaluation metrics:
\begin{itemize}
    \item \textbf{Success rate}: proportion of perturbation samples satisfying the predefined Top-30\% directional ranking criterion for the perturbed TF;
    \item \textbf{Directional accuracy}: proportion of samples in which inferred activity sign matched perturbation direction, with sign determined relative to untreated controls;
    \item \textbf{Median rank}: median rank position of the perturbed TF across evaluated samples.
\end{itemize}

\subsection*{S1.6 Pathway inference models and training settings}

Pathway classifiers used stratified 70/15/15 train/validation/test splits shared across feature configurations within each run. Seven seeds per model were run for RF, MLP, KAN, and ResNet models; the distributions and rescue counts are descriptive.

\textbf{Random forest (RF).}
RF used \texttt{n\_estimators=300}, \texttt{n\_jobs=-1}, and \texttt{random\_state} set per seed. Default \texttt{scikit-learn} settings were retained for other parameters (including \texttt{criterion=gini}, \texttt{max\_depth=None}, \texttt{min\_samples\_split=2}, \texttt{min\_samples\_leaf=1}, \texttt{max\_features=sqrt}, and \texttt{bootstrap=True}). Standard scaling was applied as part of the shared preprocessing pipeline.

\textbf{Multilayer perceptron (MLP).}
MLP comprised a 256-unit hidden layer with ReLU activation and an 11-class linear output layer. Training used AdamW (learning rate $10^{-3}$), batch size 64, a maximum of 60 epochs, and early-stopping patience of 12 epochs.

\textbf{Kolmogorov--Arnold network (KAN).}
KAN comprised two KAN layers with 128 hidden units and an 11-class output. Training used AdamW (learning rate $10^{-3}$), batch size 64, a maximum of 60 epochs, and patience of 12 epochs.

\textbf{ResNet classifier.}
The pathway ResNet classifier used a fully connected residual architecture with hidden dimension 256, three residual blocks, and dropout 0.3. Training used AdamW (learning rate $10^{-4}$, weight decay $10^{-3}$), batch size 64, a maximum of 160 epochs, early-stopping patience of 24 epochs, and cross-entropy loss. Regulatory feature layers were concatenated according to the four feature configurations defined in the main Methods.

\subsection*{S1.7 Checkpointing and evaluation protocol for pathway inference}

For MLP, KAN, and ResNet, validation macro-F1 was monitored after every epoch. The checkpoint with the highest validation macro-F1 was retained, reloaded, and then evaluated once on the corresponding held-out test partition. Training stopped after the model-specific patience interval without improvement or at the maximum epoch count.

Pathway performance metrics were overall accuracy and macro-F1.

\subsection*{S1.8 Supplementary feature-space control and recoverability analyses}

Supplementary Figure S1 provides held-out accuracy for RF, MLP, KAN, and ResNet (panel a) and macro-F1 for KAN and MLP (panel b). Both panels compare Gene, Gene augmented with PCA, AE, or RP, HubmiR alone, and Gene+HubmiR. Gene and Gene+HubmiR values are identical to the corresponding seven-run results in Figure 4. KAN and MLP share seven paired splits across all six configurations. The AE validation loss reached its minimum at epoch 54 (panel c). Per-miRNA recovery from full Gene and the 977-gene HubmiRNet input was closely aligned across all 414 outputs; 22.5\% and 23.2\% of miRNAs, respectively, exceeded a median held-out $R^2$ of 0.5 (panel d). The three aggregate $R^2$ summaries were also similar between the two source spaces and remained below 0.5 (panel e), supporting the partial linear recoverability of the HubmiR representation.

\subsubsection*{S1.8.1 Regularized canonical correlation analysis (rCCA)}

Let $X$ and $Y$ denote sample-aligned Gene and target-representation matrices. All standardization and PCA transformations were fitted on the training partition. For rCCA, let $X_{\mathrm{tr}}$ and $Y_{\mathrm{tr}}$ be their centred training scores after separate reduction to $d=64$ principal components. With $N$ training samples, define
\[
S_{XX}=\frac{X_{\mathrm{tr}}^\top X_{\mathrm{tr}}}{N-1},\quad
S_{YY}=\frac{Y_{\mathrm{tr}}^\top Y_{\mathrm{tr}}}{N-1},\quad
S_{XY}=\frac{X_{\mathrm{tr}}^\top Y_{\mathrm{tr}}}{N-1}.
\]
The regularization is scaled to the mean retained-PC variance:
\[
\lambda_X=0.1\,\frac{\operatorname{tr}(S_{XX})}{d},\qquad
\lambda_Y=0.1\,\frac{\operatorname{tr}(S_{YY})}{d}.
\]
Write $A=(S_{XX}+\lambda_X I)^{-1/2}$ and $B=(S_{YY}+\lambda_Y I)^{-1/2}$. The singular-value decomposition $A S_{XY}B=U\Sigma V^\top$ gives canonical directions $a_k=A u_k$ and $b_k=B v_k$. The reported statistic is the held-out Pearson correlation
\[
\rho_k^{\mathrm{test}}=
\operatorname{corr}(X_{\mathrm{te}}a_k,\,Y_{\mathrm{te}}b_k),
\qquad k=1,\ldots,10,
\]
rather than the training singular value. Correlations were summarized across the 20 frozen splits.

\subsubsection*{S1.8.2 Centred-kernel alignment (CKA)}

CKA used the full standardized held-out representations, without PCA reduction. For $n$ test samples, the linear and RBF kernels were
\[
K_{ij}^{\mathrm{lin}}=x_i^\top x_j,\qquad
K_{ij}^{\mathrm{RBF}}=
\exp\!\left(-\frac{\lVert x_i-x_j\rVert^2}{2s_X^2}\right),
\]
with analogous kernels $L$ for $Y$. Here $s_X^2$ and $s_Y^2$ are the respective median positive squared pairwise distances in the training partition. To reduce finite-sample bias, we used unbiased HSIC. For $K_0$ and $L_0$ obtained by setting the kernel diagonals to zero, define
\[
\begin{aligned}
h(K,L)=\frac{1}{n(n-3)}\bigg[&\operatorname{tr}(K_0L_0)\\
&+\frac{(\mathbf{1}^\top K_0\mathbf{1})(\mathbf{1}^\top L_0\mathbf{1})}{(n-1)(n-2)}\\
&-\frac{2\mathbf{1}^\top K_0L_0\mathbf{1}}{n-2}\bigg],
\end{aligned}
\]
where $\mathbf{1}$ is the all-ones vector. The reported debiased alignment was
\[
\operatorname{CKA}_{\mathrm{deb}}(K,L)=
\frac{h(K,L)}{\sqrt{h(K,K)h(L,L)}}.
\]
This estimates shared sample geometry; finite-sample estimates can be negative and do not establish molecular independence.

\subsubsection*{S1.8.3 Linear HubmiR recoverability}

For each split, let $X_{\mathrm{tr}}$ contain either all 29,045 genes or the 977 HubmiRNet inputs, and let $M_{\mathrm{tr}}$ contain the 414 inferred HubmiRs. After training-only standardization of both matrices, multi-output ridge regression solves
\[
\widehat W_\alpha=\underset{W}{\arg\min}\,
\left\{\lVert M_{\mathrm{tr}}-X_{\mathrm{tr}}W\rVert_F^2
+\alpha\lVert W\rVert_F^2\right\}.
\]
The penalty $\alpha\in\{0.01,0.1,1,10,100,1000\}$ was selected by the variance-weighted $R^2$ of pooled fivefold out-of-fold predictions within the training partition, with scalers fitted separately in each inner fold. Predictions $X_{\mathrm{te}}\widehat W_\alpha$ were transformed back to the original HubmiR scale. For test-set output $j$, define
\[
\begin{aligned}
E_j&=\sum_{i=1}^{n}(m_{ij}-\widehat m_{ij})^2,\\
T_j&=\sum_{i=1}^{n}(m_{ij}-\overline m_j)^2,\qquad
R_j^2=1-\frac{E_j}{T_j},
\end{aligned}
\]
where $\overline m_j$ is the test-set mean. With $q=414$, the aggregate summaries were
\[
\begin{aligned}
R_{\mathrm{vw}}^2&=1-\frac{\sum_j E_j}{\sum_j T_j},\\
R_{\mathrm{uniform}}^2&=\frac{1}{q}\sum_j R_j^2,\qquad
R_{\mathrm{median}}^2=\operatorname{median}_j R_j^2.
\end{aligned}
\]
Aggregate scores were summarized as mean $\pm$ s.d. across 20 splits; per-miRNA scores were summarized by their median across splits. These scores measure linear recovery of inferred features, not agreement with experimentally measured miRNA.

\subsection*{S1.9 Pathway rescue atlas and supplementary case analysis}

The rescue atlas was constructed from the frozen held-out predictions of the seven paired Gene-only and Gene+HubmiR ResNet runs used in Figure~4. Each of the 595 held-out sample--run occurrences was assigned to one of four mutually exclusive outcome states: both models incorrect, rescued (Gene-only incorrect and Gene+HubmiR correct), harmed (Gene-only correct and Gene+HubmiR incorrect), or both models correct. The respective counts were 132 (22.2\%), 50 (8.4\%), 12 (2.0\%), and 401 (67.4\%). Thus, rescues outnumbered harms by 4.17-fold, corresponding to a net gain of 38 correctly reclassified occurrences (Supplementary Figure S2a).

Rescue and harm counts were next aggregated according to the true pathway label. Rescues exceeded harms in each of the 11 pathway classes, with rescued/harmed occurrence counts of 13/6 for EGFR, 6/1 for MAPK, 7/1 for hypoxia, 5/1 for JAK--STAT, 2/0 for NF$\kappa$B, 1/0 for PI3K, 2/1 for TGF$\beta$, 7/1 for TNF$\alpha$, 1/0 for TRAIL, 3/0 for VEGF, and 3/1 for p53 (Supplementary Figure S2b).

For the profile-level catalogue, the nine EGFR, MAPK, and hypoxia cases displayed in Figure~6 were retained. For each of the other eight pathways, the run containing the largest number of distinct strict rescues was identified first, and profiles from that run were prioritized. Remaining positions were filled in descending order of rescue-minus-harm stability across the seven runs. A maximum of three profiles was retained per pathway, and pathways with fewer than three distinct strict rescues were not supplemented with non-rescue cases. The final catalogue therefore contained 27 strict-rescue profiles spanning all 11 pathways: three each for EGFR, MAPK, hypoxia, JAK--STAT, TNF$\alpha$, VEGF, and p53; two each for NF$\kappa$B and TGF$\beta$; and one each for PI3K and TRAIL (Supplementary Figure S2c).

Supplementary Figure S3 displays eight selected genes and three inferred HubmiRs for each pathway across the same 27 profiles. As in Figure~6, feature values were converted to training-fold empirical-percentile deviations, $d=2\{F_{\mathrm{train}}(x)-0.5\}$, using the feature-specific empirical cumulative distribution $F_{\mathrm{train}}$ estimated only from the corresponding outer-training fold. The scale therefore ranges from $-1$ to $+1$, with zero denoting the training-fold median. For profiles rescued in multiple runs, the plotted value is the mean across those strict-rescue occurrences.

For EGFR, MAPK, and hypoxia, the supplementary atlas preserves the genes and inferred HubmiRs used in Figure~6. For the remaining pathways, the panels were assembled from pathway-specific features across the frozen rescue cases. The p53 panel required a separate footprint audit because the available official p53 response genes remained strongly shifted in at least one selected profile. It therefore combines four official p53 footprint genes (\textit{PLTP}, \textit{CABYR}, \textit{RETSAT}, and \textit{ARHGAP11A}) with four low-intensity context genes (\textit{MTMR10}, \textit{ZXDB}, \textit{EPS8}, and \textit{TPSG1}); these roles are identified in Supplementary Figure S3. Feature selection was confined to visualization and did not alter classifier training, prediction, or the rescue/harm definitions.

\clearpage
\begin{figure*}[p]
    \centering
    \includegraphics[width=\textwidth,height=0.78\textheight,keepaspectratio]{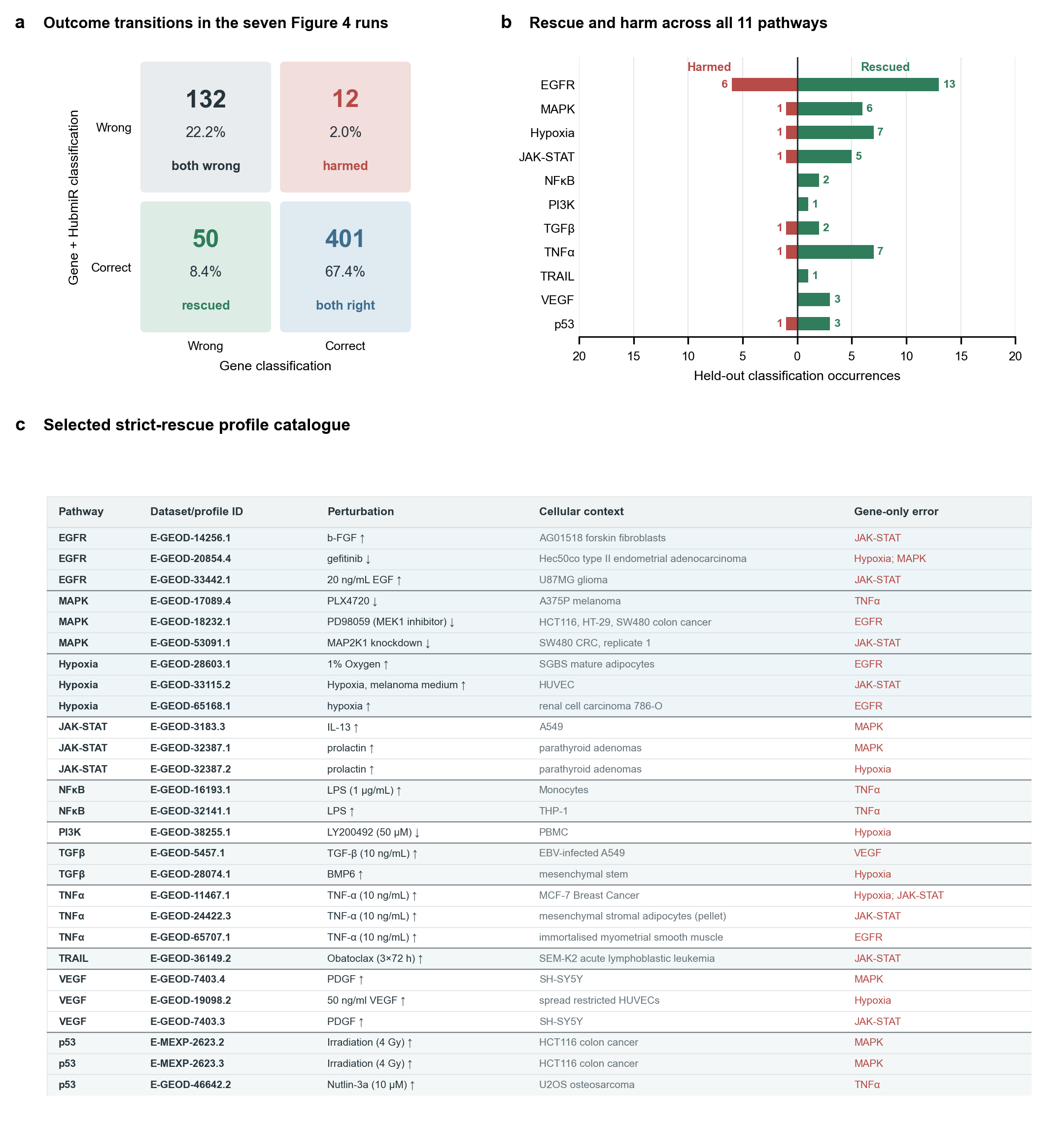}
    \caption[Pathway-wide rescue outcomes and selected strict-rescue catalogue across seven ResNet runs.]{\textbf{Pathway-wide rescue outcomes and selected strict-rescue catalogue across seven ResNet runs.} \textbf{a,} Joint correctness of the paired Gene-only and Gene+HubmiR predictions across 595 held-out sample--run occurrences. Rescued denotes Gene-only incorrect and Gene+HubmiR correct; harmed denotes Gene-only correct and Gene+HubmiR incorrect. Counts and percentages use all 595 paired occurrences as the denominator. \textbf{b,} Rescued and harmed occurrence counts stratified by the true pathway label. \textbf{c,} Catalogue of the 27 selected strict-rescue profiles displayed in Supplementary Figure S3, annotated by pathway, dataset/profile identifier, perturbation, cellular context, and the incorrect Gene-only prediction. At most three profiles were selected per pathway; pathways with fewer available strict rescues were not padded. The nine profiles examined in Figure~6 were retained in the catalogue.}
    \label{fig:supp_rescue_atlas}
\end{figure*}

\clearpage
\begin{figure*}[p]
    \centering
    \includegraphics[width=\textwidth,height=0.87\textheight,keepaspectratio]{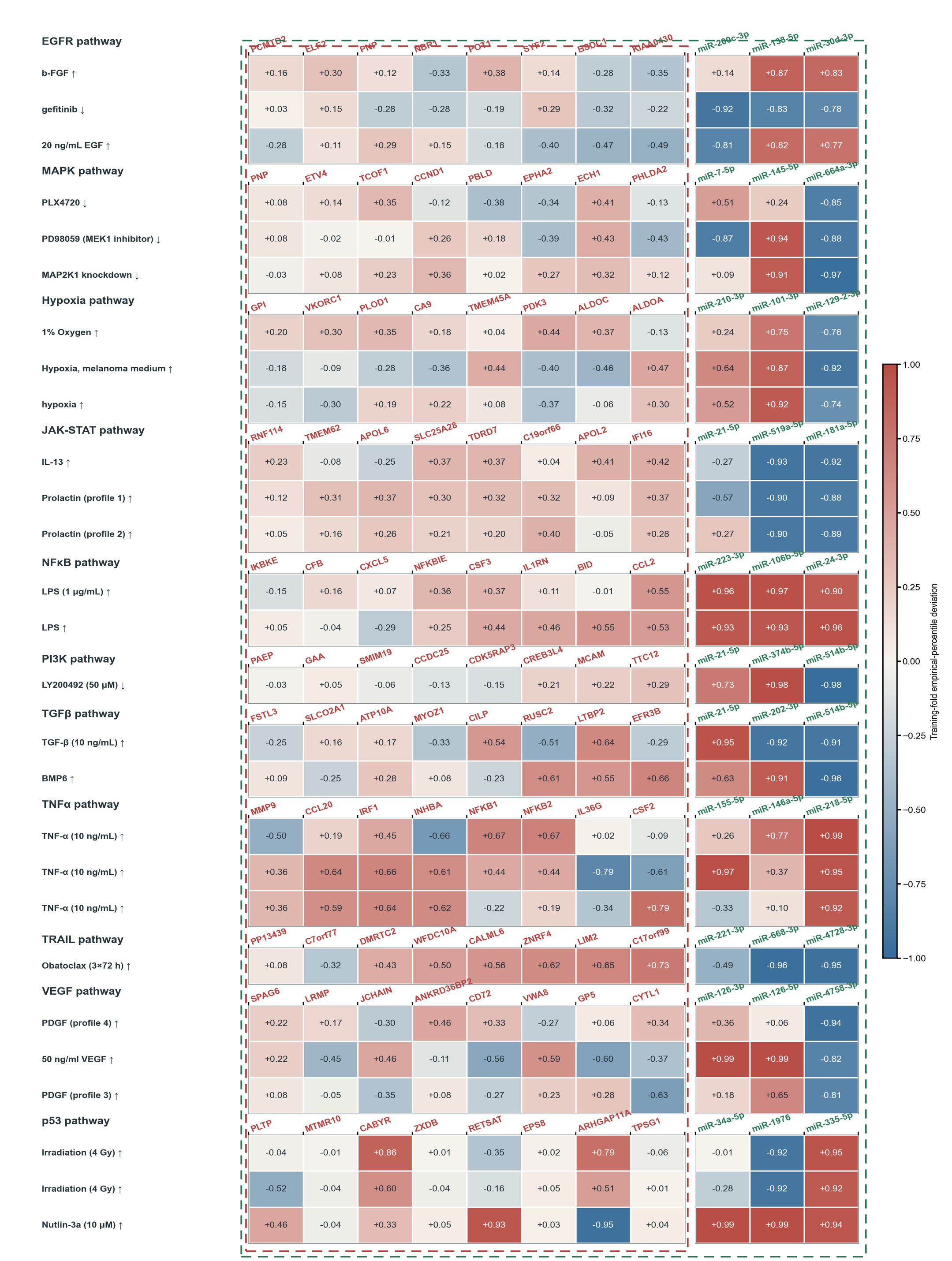}
    \caption[Feature profiles of selected strict-rescue cases across 11 pathways.]{\textbf{Feature profiles of selected strict-rescue cases across 11 pathways.} Fold-relative empirical-percentile deviations for eight selected genes and three inferred HubmiRs across the same 27 strict-rescue profiles catalogued in Supplementary Figure S2c. Red dashed frames denote the Gene-only features, whereas green dashed frames encompass the genes together with the inferred HubmiRs available to the combined classifier. Values range from $-1$ to $+1$, with zero denoting the corresponding outer-training-fold median; values for profiles rescued in multiple runs are averaged across their strict-rescue occurrences.}
    \label{fig:supp_rescue_features}
\end{figure*}

\clearpage

\section*{S2. Mathematical definitions of drug MoA similarity algorithms}

To quantify similarity between drug-induced transcriptional signatures, we evaluated four widely used connectivity scoring methods: Kolmogorov--Smirnov (KS), weighted KS, extreme cosine similarity (XCos), and ZhangScore. These methods capture concordance between a query signature and a reference signature from complementary perspectives, including enrichment of extreme genes, magnitude-weighted enrichment, cosine similarity of extreme components, and signed rank agreement.

Let
\[
\mathbf{x} = (x_1,\dots,x_n), \qquad \mathbf{y} = (y_1,\dots,y_n)
\]
denote the query and reference signatures, respectively, defined on the same set of $n$ genes. Each entry represents a signed differential-expression statistic, such as a moderated $t$-statistic, $z$-score, or log fold-change. Positive values indicate up-regulation and negative values indicate down-regulation.

For the query signature $\mathbf{x}$, we define the sets of most strongly up- and down-regulated genes as
\[
U(\mathbf{x}) = \mathrm{Top}_k(\mathbf{x}), \qquad D(\mathbf{x}) = \mathrm{Bottom}_k(\mathbf{x}),
\]
where $\mathrm{Top}_k(\mathbf{x})$ and $\mathrm{Bottom}_k(\mathbf{x})$ denote the indices of the $k$ largest and $k$ smallest entries of $\mathbf{x}$, respectively. Throughout, all signatures were assumed to be pre-aligned by gene identity.

For rank-based methods, let
\[
\pi_{\mathbf{y}} = (\pi_1,\dots,\pi_n)
\]
be the permutation of gene indices such that
\[
y_{\pi_1} \ge y_{\pi_2} \ge \cdots \ge y_{\pi_n}.
\]
Accordingly, genes near the top of the ordered list are the most up-regulated in the reference signature, whereas genes near the bottom are the most down-regulated.

\subsection*{S2.1 Kolmogorov--Smirnov (KS) score}

The KS score follows the original Connectivity Map paradigm and evaluates whether the query up- and down-regulated genes are preferentially enriched at opposite ends of the ranked reference signature.

For a gene set $S \subseteq \{1,\dots,n\}$ with cardinality $|S| = N_S$, define the indicator
\[
I_i(S) =
\begin{cases}
1, & \pi_i \in S,\\
0, & \pi_i \notin S.
\end{cases}
\]
The running-sum statistic over the ranked reference profile is
\[
RS_j(S;\mathbf{y})
=
\sum_{i=1}^{j}
\left(
\frac{I_i(S)}{N_S}
-
\frac{1-I_i(S)}{n-N_S}
\right),
\qquad j=1,\dots,n.
\]
The enrichment score of $S$ with respect to $\mathbf{y}$ is defined as the extremum with the largest absolute deviation:
\[\resizebox{0.98\columnwidth}{!}{$\displaystyle
ES(S;\mathbf{y})
=
\begin{cases}
\max_j RS_j(S;\mathbf{y}), & \text{if } \max_j RS_j(S;\mathbf{y}) \ge -\min_j RS_j(S;\mathbf{y}),\\[4pt]
\min_j RS_j(S;\mathbf{y}), & \text{otherwise.}
\end{cases}
$}\]
The KS connectivity score between $\mathbf{x}$ and $\mathbf{y}$ is then
\[
\mathrm{KS}(\mathbf{x},\mathbf{y})
=
ES\!\left(U(\mathbf{x});\mathbf{y}\right)
-
ES\!\left(D(\mathbf{x});\mathbf{y}\right).
\]
A large positive value indicates that genes up-regulated in the query tend to be highly ranked in the reference signature, while genes down-regulated in the query tend to be enriched near the bottom, corresponding to a concordant transcriptional response.

\subsection*{S2.2 Weighted KS score}

The weighted KS score extends the unweighted KS formulation by assigning greater influence to genes with larger absolute differential-expression values in the reference signature. Let
\[
w_i = |y_{\pi_i}|^p,
\]
where $p > 0$ is the weighting exponent. In this study, we used $p=1$, corresponding to the standard weighted enrichment formulation.

For a gene set $S$, define the normalization constant
\[
W_S = \sum_{i:\,\pi_i \in S} w_i.
\]
The weighted running-sum statistic is
\[
RS^{(p)}_j(S;\mathbf{y})
=
\sum_{i=1}^{j}
\left(
\frac{I_i(S)\,w_i}{W_S}
-
\frac{1-I_i(S)}{n-N_S}
\right),
\qquad j=1,\dots,n.
\]
The corresponding weighted enrichment score is
\[\resizebox{0.98\columnwidth}{!}{$\displaystyle
ES^{(p)}(S;\mathbf{y})
=
\begin{cases}
\max_j RS^{(p)}_j(S;\mathbf{y}), & \text{if } \max_j RS^{(p)}_j(S;\mathbf{y}) \ge -\min_j RS^{(p)}_j(S;\mathbf{y}),\\[4pt]
\min_j RS^{(p)}_j(S;\mathbf{y}), & \text{otherwise.}
\end{cases}
$}\]
The weighted KS connectivity score is defined as
\[
\mathrm{weightedKS}(\mathbf{x},\mathbf{y})
=
ES^{(1)}\!\left(U(\mathbf{x});\mathbf{y}\right)
-
ES^{(1)}\!\left(D(\mathbf{x});\mathbf{y}\right).
\]
Relative to the unweighted KS score, this variant places greater emphasis on genes exhibiting stronger transcriptional perturbation in the reference profile.

\subsection*{S2.3 Extreme cosine similarity (XCos)}

XCos measures similarity by restricting attention to the most extreme components of the two signatures and then computing cosine similarity in that reduced space.

Let
\[
T_k(\mathbf{x}) = U(\mathbf{x}), \qquad B_k(\mathbf{x}) = D(\mathbf{x}),
\]
and define the corresponding binary masks
\[
m_i^{\mathrm{top}}(\mathbf{x}) =
\begin{cases}
1, & i \in T_k(\mathbf{x}),\\
0, & \text{otherwise,}
\end{cases}
\qquad
m_i^{\mathrm{bot}}(\mathbf{x}) =
\begin{cases}
1, & i \in B_k(\mathbf{x}),\\
0, & \text{otherwise.}
\end{cases}
\]
We then construct an extreme-signature representation by concatenating the top and bottom components:
\[\resizebox{0.98\columnwidth}{!}{$\displaystyle
\phi(\mathbf{x})
=
\Bigl(
x_1 m_1^{\mathrm{top}}(\mathbf{x}),\dots,x_n m_n^{\mathrm{top}}(\mathbf{x}),
x_1 m_1^{\mathrm{bot}}(\mathbf{x}),\dots,x_n m_n^{\mathrm{bot}}(\mathbf{x})
\Bigr)\in\mathbb{R}^{2n}.
$}\]
The same construction is applied to the reference signature to obtain $\phi(\mathbf{y})$. The XCos score is defined as
\[
\mathrm{XCos}(\mathbf{x},\mathbf{y})
=
\frac{\phi(\mathbf{x})^\top \phi(\mathbf{y})}
{\|\phi(\mathbf{x})\|_2\,\|\phi(\mathbf{y})\|_2}.
\]
If either denominator is zero, the score is set to zero. By construction, XCos suppresses the contribution of weakly perturbed genes and focuses on agreement among the most transcriptionally extreme components.

\subsection*{S2.4 ZhangScore}

ZhangScore quantifies concordance using the signed rank positions of genes in the reference signature. Let $r_i(\mathbf{y}) \in \{1,\dots,n\}$ denote the rank of gene $i$ in the reference profile, with larger rank corresponding to stronger up-regulation. These ranks are transformed to signed normalized ranks:
\[
s_i(\mathbf{y}) = \frac{2\,r_i(\mathbf{y})}{n} - 1,
\]
so that $s_i(\mathbf{y}) \in [-1,1]$, with positive values assigned to genes near the top of the ranked list and negative values assigned to genes near the bottom.

The mean signed rank of the query up- and down-regulated genes is then given by
\[
\mu_U(\mathbf{x},\mathbf{y})
=
\frac{1}{|U(\mathbf{x})|}
\sum_{i\in U(\mathbf{x})} s_i(\mathbf{y}),
\qquad
\mu_D(\mathbf{x},\mathbf{y})
=
\frac{1}{|D(\mathbf{x})|}
\sum_{i\in D(\mathbf{x})} s_i(\mathbf{y}).
\]
The ZhangScore is defined as
\[
\mathrm{ZhangScore}(\mathbf{x},\mathbf{y})
=
\mu_U(\mathbf{x},\mathbf{y}) - \mu_D(\mathbf{x},\mathbf{y}).
\]
Positive values indicate that genes up-regulated in the query tend to occupy highly ranked positions in the reference signature, whereas genes down-regulated in the query tend to occupy lowly ranked positions, consistent with transcriptional concordance.

\subsection*{S2.5 Interpretation}

For all four methods, larger positive scores indicate stronger similarity between the query and reference transcriptional responses, whereas negative scores indicate an inverse or reversal-like relationship. Because the numerical ranges and scaling properties differ across algorithms, score magnitudes are most appropriately compared within, rather than across, methods.

\subsection*{S2.6 Practical implementation}

In our implementation, all methods were computed on gene-aligned signature vectors. KS and weighted KS used the extreme up- and down-regulated genes of the query signature to define directional gene sets; XCos was computed from the extreme components of both signatures; and ZhangScore was based on the signed rank positions of the query extreme genes in the ranked reference profile.

\end{document}